\documentclass{article}

     \PassOptionsToPackage{numbers, compress}{natbib}
\usepackage[preprint]{neurips_2025}

\usepackage[utf8]{inputenc} % allow utf-8 input
\usepackage[T1]{fontenc}    % use 8-bit T1 fonts
\usepackage{hyperref}       % hyperlinks
\usepackage{url}            % simple URL typesetting
\usepackage{booktabs}       % professional-quality tables
\usepackage{amsfonts}       % blackboard math symbols
\usepackage{nicefrac}       % compact symbols for 1/2, etc.
\usepackage{microtype}      % microtypography
\usepackage{xcolor}         % colors
\usepackage{graphicx}
\usepackage{multirow}
\usepackage{amsmath}
\usepackage[numbers]{natbib}
\usepackage{wrapfig}

\title{Benchmarking Post-Training Quantization in LLMs: Comprehensive Taxonomy, Unified Evaluation, and Comparative Analysis}

\author{Jiaqi Zhao\textsuperscript{1}\thanks{Equal Contribution}\hspace{0.5em}, Ming Wang\textsuperscript{1}\footnotemark[1]\hspace{0.5em}, Miao Zhang\textsuperscript{1}\thanks{Corresponding Author}\hspace{0.5em}, Yuzhang Shang\textsuperscript{2}\footnotemark[2]\hspace{0.5em}, Xuebo Liu\textsuperscript{1}, Yaowei Wang\textsuperscript{1}, \\
        \bf{Min Zhang\textsuperscript{1}, Liqiang Nie\textsuperscript{1}}\\
        \textsuperscript{1} Harbin Institute of Technology (Shenzhen)\\
        \textsuperscript{2} Illinois Institute of Technology\\
        \texttt{jiaqizhao0455@outlook.com, 190110509@stu.hit.edu.cn, yshang4@hawk.iit.edu}\\
        \texttt{\{zhangmiao, liuxuebo, wangyaowei, zhangmin2021\}@hit.edu.cn}, nieliqiang@gmail.com}
\begin{document}

\maketitle

\begin{abstract}
Post-training Quantization (PTQ) technique has been extensively adopted for large language models (LLMs) compression. However, existing research lacks in-depth analysis of the strengths and applicable scenarios of different PTQ strategies, making it difficult for future researchers to choose suitable foundational framework for development based on their specific needs. To mitigate these confusions, we start by proposing a detailed taxonomy for existing mainstream methods based on their computational strategies (e.g., optimization-based, compensation-based, etc.). Subsequently, we establish a comprehensive benchmark for LLMs PTQ, named PTQ-Bench, to help select foundational framework, which evaluates the cross-bitwidth robustness, cross-structure robustness and cross-modality robustness of PTQ frameworks. Using PTQ-Bench, we conduct extensive experiments with the baseline of each categorized PTQ strategy, covering models with various sizes (7B-70B), bitwidths, structures (LLaMA1/2/3/3.1, Mixtral, DeepSeekMoE and Mamba) and modality (LLaVA1.5 and VILA1.5) on a wide range of evaluation metrics, followed by comparative analysis on the results to summarize the superior of each PTQ strategy. For example, our benchmark reveals that rotation-based technique demonstrates outstanding low-bit robustness. Finally, beyond benchmarking we present two critical implications: (1) extremely low-bit PTQ for ultra large models should be reexamined, and (2) a practical combination of compensation and other PTQ strategy can achieve SOTA various robustness. We believe our PTQ-Bench will provide valuable recommendations for future research on advanced PTQ methods. The repository for our PTQ-Bench is available at \url{https://github.com/zjq0455/PTQ-Bench}.
\end{abstract}

\section{Introduction}

\begin{figure*}[t]
\centering
\includegraphics[width=5.5in]{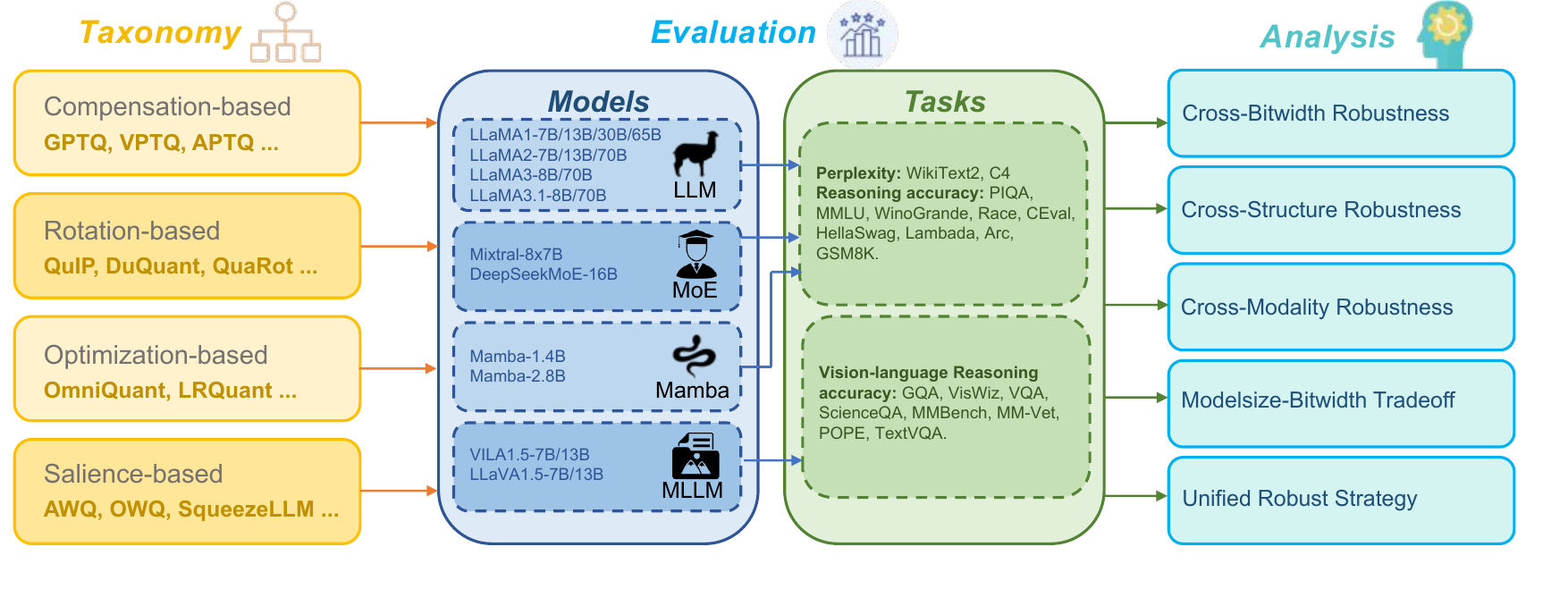}
\caption{An overview of our paper. To provide guidelines for future research, we first establish a comprehensive taxonomy for existing milestone PTQ methods. Then we establish a novel benchmark named PTQ-Bench for evaluating several critical characteristics of foundational PTQ strategies. Based on it, extensive and unified evaluation of the categorized PTQ strategies is provided which contains a broad range of model sizes, structures, modalities and bitwidth. Finally, we summarize in-depth comparative analysis based on the experimental results and offer valuable recommendations for the advancement of LLM PTQ research.}
\label{overview}
\end{figure*}

Large language models (LLMs) have achieved remarkable success in text generation and various reasoning tasks, with representative ChatGPT \citep{achiam2023gpt} and LLaMA family \citep{touvron2023llama1, touvron2023llama2,dubey2024llama}. However, their massive parameter scale imposes significant memory and inference overhead, which constrain their practical deployment. To address the issue, numerous model compression techniques have been proposed, such as quantization \citep{lee2024owq,shang2024enhancing}, pruning \citep{frantar2023sparsegpt,cheng2024influence,cheng2024survey}, low-rank decomposition \citep{hu2021lora,yuan2023asvd}, and knowledge-distillation \citep{gou2021knowledge}. Among these, Post-training Quantization (PTQ) \citep{yao2022zeroquant,li2023fptq}, a technique unlike Quantization-aware Training (QAT) \citep{liu2023llm, wang2023bitnet,xu2024onebit} which requires heavily retraining, has been widely employed due to its efficiency and resource-friendly nature. As illustrated by Figure \ref{trend}, the number of papers about PTQ for LLMs takes up nearly 70\% of total quantization papers since 2022. 

Despite the growing prominence of PTQ, current research still exhibits the following limitation: the present reviews lack of in-depth insights into the characteristic of different PTQ frameworks so as to provide limited guidance for the development on advanced PTQ methods. These reviews either exhibit a lack of adequate focus on quantization \citep{tang2024survey,yang2024llmcbench}, incorporate insufficient experimental setups \citep{gong2024survey,kurtic2024give}, or offer inadequate analytical insights \citep{li2024evaluating}. Admittedly, it would hold greater value for future researchers to get the information on selecting foundational PTQ frameworks for further exploration on their specific scenarios. For example, a typical question might be: \textbf{\textit{Which foundational PTQ framework should I choose to achieve better robustness across various model structure?}}

% \textbf{Secondly}, existing PTQ methods invariably prioritize enhancing the quantization performance while overlooking the trade-off among performance, model size, and quantization bitwidth. For instance, considering a intuitive questions: \textbf{\textit{Which one is better, a higher-bit small model or a lower-bit large model?}} This is directly relevant to the selection of quantized LLMs for deployment, but previous studies cannot answer this question because they merely applied the proposed methods to models of various sizes for evaluation and presented experimental results to demonstrate their superiority without delving into the specifics.

To fill this confusion, in this paper we introduce a novel benchmark for foundational PTQ framework selection named PTQ-Bench. As illustrated by Figure \ref{overview}, our benchmark is built upon a comprehensive taxonomy of PTQ methods, unified evaluation including extensive experiments, and comparative analysis on the results to offer valuable recommendations. Particularly, we make full attention on weight-only PTQ methods, as they have been adopted in a greater number of practical applications, enables a wider range of bitwidth support, exhibits a richer diversity of strategies and demonstrates superior performance compared to weight-activation methods \citep{yuan2024llm}. Our contributions are revealed in four dimensions:

\textbf{(1) Comprehensive Taxonomy.} We first review the vast majority of mainstream weight-only PTQ techniques and categorize them into four classes based on their designing strategies and optimization mechanisms: \textbf{compensation-based strategy} represented by GPTQ \citep{frantar2022gptq}, \textbf{optimization-based strategy} exemplified by OmniQuant \citep{shao2023omniquant}, \textbf{rotation-based strategy} typified by QuIP \citep{chee2024quip}, and \textbf{salience-based strategy} characterized by AWQ \citep{lin2024awq}. Our proposed taxonomy can provide a clear understanding for researchers and support our subsequent benchmarking experiments.

% \textbf{(2) Unified Evaluation.} In order to derive the characteristics of each PTQ strategy and provide reasonable and accurate recommendations, we establish a comprehensive benchmark by extensively evaluate their cross bitwidth/training-level robustness, cross-structure robustness and cross modality robustness 

\textbf{(2) PTQ-Bench for Unified Evaluation.} To provide guidelines for facilitating the development of more advanced PTQ methods, we establish a comprehensive benchmark, namely PTQ-Bench, by evaluating the \textbf{cross-bitwidth robustness}, \textbf{cross-structure robustness} and \textbf{cross-modality robustness} of each PTQ strategy through extensive experiments to derive their characteristics. 
%, and introducing the \textbf{modelsize-bitwidth trade-off} by comparing the performance of models across various sizes at different bitwidth. 
In detail, the experiments are mainly constructed on the most widely used open-sourced LLMs LLaMA family (LLaMA-1/2/3/3.1) with a large range of model size (7B to 70B). For comprehensiveness and unification, Mixture-of-Experts (MoE) LLMs, Mamba \citep{gu2023mamba} and multimodal LLMs (MLLMs) are also included. Extremely low-bit quantization (2-bit) to common 4-bit quantization are applied to all LLMs using representative baselines from the each PTQ strategy and various evaluation tasks are covered such as language modeling and reasoning.

\textbf{(3) Comparative Analysis.} Based on the unified evaluation, we summarize the characteristics of each PTQ strategy in different scenarios, thus offering practical recommendations for the future researchers to select foundational PTQ frameworks based on their requirements. For example, we suggest to select compensation-based strategy when facing cross-structure scenario and avoid choosing naive optimization-based strategy at extremely low-bit quantization. 

\textbf{(4) Beyond Benchmarking.} We also present the following two critical conclusion in LLMs PTQ deriving from PTQ-bench: a) Even the largest model at 2-bit performs worse than the smallest model at 4-bit in the same LLM family, so we accordingly argue that the extremely low-bit PTQ for ultra large models need to be reexamined; b) We innovatively claim that practical combination of compensation-based scheme and other PTQ strategy can achieve SOTA various robustness.

\section{Taxonomy}
\label{section:taxonomy}

Most previous reviews have also categorized PTQ techniques, such as symmetric or asymmetric \citep{gholami2022survey} and group-wise or channel-wise quantization \citep{shen2020q}. Such taxonomies have become increasingly inadequate to meet the research requirements of subsequent studies as PTQ techniques continue to proliferate, because such taxonomies are relatively coarse-grained, making them challenging to conduct in-depth analysis of the characteristics of each category. In this section, we compile a comprehensive list of existing mainstream weight-only PTQ algorithms and categorize them based on their underlying principles. Specifically, they are classified into four categories: \textbf{compensation-based strategy}, \textbf{rotation-based strategy}, \textbf{salience-based strategy} and \textbf{optimization-based strategy}.

\subsection{Compensation-based Quantization}

The strategy of compensation-based technique is to dynamically update the weights to compensate for quantization errors during the process. Specifically, these methods typically calculate the bad impact of quantization and then derive the required compensation in order to mitigate this impact. 

This strategy is pioneered by GPTQ \citep{frantar2022gptq} and is one of the most influential quantization techniques currently. GPTQ first reformulates the quantization error involving the Hessian matrix and then calculates the update formula for the unquantized weights after a specific weight is quantized, which can be expressed as:
\begin{equation}
\delta = -\frac{w_q - \text{quant}(w_{q})}{[\mathbf{H}^{-1}]_{qq}}\cdot(\mathbf{H}^{-1})_{:,q},
\end{equation}
where $\delta$ denotes the optimal update of the unquantized weights, $w_{q}$ is the weight as position $q$ and $\mathbf{H}$ indicates Hessian. Unlike \citet{lecun1989optimal} and \citet{frantar2022optimal} which require a greedy search to identify the position that minimizes the errors for each quantization step, GPTQ partitions the weight matrix into multiple blocks and performs column-wise operations sequentially, during which the residual weights within the current block are compensated accordingly. Following GPTQ, more advanced error compensation strategies are proposed by QuantEase \citep{behdin2023quantease}, VPTQ \citep{liu2024vptq} and APTQ \citep{guan2024aptq}, achieving nearly no performance degradation even at 3-bit quantization.

\subsection{Rotation-based Quantization}

The development of rotation-based methods stems from the observation that the distribution of pretrained weights in LLMs does not facilitate the direct quantification, such as the existence of outliers. Targeting this issue, researchers typically apply transformations to process the weight matrix to enhance quantization performance. 

QuIP \citep{chee2024quip} is considered as the innovator of this strategy. Their insights reveal that quantization will be more effective when weights and proxy Hessian are incoherent, where a weight matrix $W \in \mathbb{R}^{n \times m}$ is $\mu$-incoherent if:
\begin{equation}
%    \resizebox{.85\linewidth}{!}{$
\max_{i,j} |W_{ij}| = \max_{i,j} \left| e_i^T W e_j \right| \leq \mu \|W\|_F / \sqrt{mn}.
%  $}
\end{equation}
Specifically, QuIP multiplies a weight matrix by Kronecker-structured orthogonal matrices \citep{zhang2015fast} on the left and right. Such process can be thought of as a principled form of outlier reduction because the weights are similar in magnitudes, ensuring the rounding errors are not particularly large in any direction along the coordinate axes. 

Inspired by QuIP, rotation-based methods rapidly gained traction. QuIP\# \citep{tseng2024quip} employs Hadamard matrices \citep{halko2011finding} for rotation, achieving more efficient and superior quantization performance. QuaRot \citep{ashkboos2024quarot} and SpinQuant \citep{liu2024spinquant} further extend the Hadamard-rotation method to weight-activation quantization to eliminate the extreme outliers in activation channels.

\subsection{Salience-based Quantization}

Salience-based strategy asserts the weights in LLMs exhibit varying degrees of importance, and quantization performance would be improved by selectively handling them based on their saliency. Generally, the motivation behind these algorithms primarily revolves around the criteria for determining salience and the treatment methods applied to the weights from different groups.

Mixed-precision quantization methods constitute the largest subset within salience-based technique. These methods retain salient weights at higher precision while quantizing the others to lower bits. LLM.int8() \citep{dettmers2022gpt3} performs 8-bit quantization while preserving the weights with the top-$0.1\%$ magnitudes at 16-bit. SpQR \citep{dettmers2023spqr} saves more unstructured salient weights at higher precision and employs a finer-grained group-wise quantization approach. PB-LLM \citep{shang2023pb} adopts Hessian which is considered as a more advanced metric to isolate salient weights for 8-bit quantization while binarizing the others.

Although yielding promising results, mixed-precision quantization is not hardware-friendly which may harms the inference speed and requires specialized design to accommodate the varying bit widths. Taking it into considerations, AWQ \citep{lin2024awq}, the representative method of salience-based technique, is proposed. Firstly, AWQ relies on input activation as the measure instead of the previous ones that use the weights themselves as the criterion for salience. Following this, they conclude that scaling the salient weights will reduce quantization errors, thereby avoiding the deployment challenges associated with mixed-precision quantization. The scaling quantization for salient weights can be elaborated as:
\begin{equation}
Q(w \cdot s) \cdot \frac{x}{\Delta} = \Delta \cdot \text{Round}(\frac{ws}{\Delta}) \cdot x \cdot \frac{1}{s},
\end{equation}
where $\Delta$ is the quantization scalar and $s$ denotes the scaling hyper-parameter. AWQ is currently one of the most widely applied PTQ method like GPTQ .

\subsection{Optimization-based Quantization}

A common characteristic of the three categories above is leveraging the intrinsic properties of the weights to influence the quantization outcome. Meanwhile, some researchers indicate that it is also effective to employ efficient optimization framework to update the quantization parameters. Since LLM weights are frozen and the optimization process is highly efficient, such approaches are also referred to as PTQ techniques

OmniQuant \citep{shao2023omniquant} is the first to introduce optimization strategy into PTQ. To avoid insufferable computational resources cost during training, an efficient block-wise learning framework is proposed, where the output of full-precision blocks serves as supervisory information to update the clipping range of the scaling factors, achieving nearly lossless performance at 4-bit quantization. The optimization objective is:
\begin{equation}
%    \resizebox{.85\linewidth}{!}{$
\mathop{\arg\min}\limits_{\Theta_{1},\Theta_{2}}\|F(\mbox{W},\mbox{X})-F(Q_{w}(\mbox{W}; \Theta),X)\|,
%$}
\end{equation}
where $F$ means the mapping function for the current block and $X$ denotes the full-precision activation. $Q_{w}(\cdot)$ represents the weight quantizer. $\Theta$ indicates learnable scaling factors. 

Following OmniQuant, CBQ \citep{ding2023cbq} devises a two-branch framework to improve the robustness. LRQuant \citep{zhao2024lrquant} discovers the directional gaps between full-precision outputs and their quantized counterparts and propose a novel loss function named NLC loss to minimize the quantization error. AffineQuant \citep{ma2024affinequant} incorporates affine transformations into the PTQ process and optimizes the transformation matrix to reduce quantization errors. %\textcolor{red}{IMPORTANT!}

Despite our definitive taxonomy of existing milestone PTQ methods, the specific performance traits and suitable application contexts of each strategy are still unclear. The future researchers are still confused by the selection of foundational PTQ framework based on their requirements, this necessitating further analysis based on experiments.

\begin{table*}[t]
%  \tiny
  \centering
    \caption{The average perplexity($\downarrow$)/accuracy($\%$) comparison results on LLaMA-1/2. For detailed results please refer to Table \ref{llama1-pfm} and \ref{llama2-pfm} in Appendix \ref{detailed}.}
  \resizebox{\textwidth}{!}{
  \begin{tabular}{c||c|cccc|cccc}
    \toprule[1.5pt]
    \textbf{Methods} &W4 & \multicolumn{4}{c|}{W3} & \multicolumn{4}{c}{W2}\\
     %\cmidrule(r){3-11} %\cmidrule(r){4-7} \cmidrule(l){9-12}
    \midrule
    \textbf{LLaMA-} &7B & 7B  & 13B & 30B & 65B & 7B & 13B & 30B & 65B\\ 
    \midrule %\cmidrule(r){4-7} \cmidrule(l){9-12}
    AWQ &\textbf{6.50/51.32} &\textbf{7.08/49.91} &\textbf{6.30/53.28} &\textbf{5.48/58.73} &\textbf{4.94/61.97} &2.7e5/22.15 &2.5e5/22.88 &2.3e5/22.98 &7.4e4/22.86\\
    GPTQ &7.07/50.19 &9.29/41.33 &6.40/51.95 &5.70/57.39 &5.10/60.85 &35.86/25.48 &16.74/30.89 &13.38/33.98 &9.53/43.44\\
    QuIP &7.11/49.16 &8.87/44.66 &6.42/52.44 &5.61/57.07 &5.22/59.47 &\textbf{20.66/31.73} &\textbf{12.60/35.97} &\textbf{10.14/40.49} &\textbf{7.90/47.54}\\
    OmniQ &6.61/51.01 &7.38/48.21 &6.51/51.28 &5.67/56.87 &5.08/59.71 &29.77/29.12 &16.13/34.16 &11.78/38.76 &9.37/42.81\\
    \cmidrule{1-10}
    \textbf{LLaMA2-} &7B & 7B  & 13B & 70B & - & 7B & 13B & 70B & -\\ 
    \midrule %\cmidrule(r){4-7} \cmidrule(l){9-12}
    AWQ &\textbf{6.36/53.05} &\textbf{7.02/50.69} &\textbf{6.14/55.30} &\textbf{4.78/64.35} &- &1.9e5/22.51 &1.1e5/22.61 &6.9e4/22.50 &-\\
    GPTQ &6.45/52.86 &7.20/49.23 &6.28/55.44 &4.88/63.19 &- &57.92/25.21 &19.50/29.55 &9.00/43.44 & -\\
    QuIP &6.75/50.44 &19.55/35.23 &6.39/53.98 &5.25/61.60 &- &\textbf{43.75/26.27} &\textbf{14.23/33.98} &\textbf{7.70/48.47} & -\\
    OmniQ &6.55/51.11 &7.62/47.57 &6.49/53.34 &5.00/62.58 &- &58.04/25.22 &22.48/30.47 &10.36/38.46 &- \\
   \bottomrule[1.5pt]
  \end{tabular}}
  \label{llm-ppl}
\vspace{-1em}
\end{table*}

\begin{table*}[t]
  \scriptsize
  \centering
  \setlength{\tabcolsep}{10pt}
  \renewcommand{\arraystretch}{0.7} 
    \caption{The average perplexity($\downarrow$)/accuracy($\%$) comparison results on LLaMA-3/3.1. For detailed results please refer to Table \ref{llama3-pfm} and \ref{llama31-pfm} in Appendix \ref{detailed}.}
  \begin{tabular}{c||c|cc|cc}
    \toprule[1.2pt]
    \textbf{Methods} &W4 & \multicolumn{2}{c|}{W3} & \multicolumn{2}{c}{W2}\\
     %\cmidrule(r){3-11} %\cmidrule(r){4-7} \cmidrule(l){9-12}
    \midrule
    \textbf{LLaMA3-} &8B &8B & 70B  &8B & 70B  \\ 
    \midrule %\cmidrule(r){4-7} \cmidrule(l){9-12}
    AWQ &7.98/\textbf{61.23} &\textbf{9.83/54.65} & \textbf{6.30/69.53} &1.9e6/22.26 &1.6e6/22.54 \\
    GPTQ &\textbf{7.96}/60.77 &10.84/52.19 &7.16/40.51 &647.68/22.82 &\textbf{23.43/29.93}\\
    QuIP &8.73/58.53 &10.09/53.20 &68.60/26.76  &\textbf{130.32/23.05} &53.18/24.60 \\
    OmniQ &8.82/57.36 &17.53/36.09 &7.4e4/22.22  &1.5e3/22.67 &4.1e4/22.30  \\
    \midrule
    \textbf{LLaMA3.1-} &8B &8B & 70B &8B & 70B \\ 
    \midrule %\cmidrule(r){4-7} \cmidrule(l){9-12}
    AWQ &\textbf{8.09/61.60} &\textbf{9.91/55.50} &\textbf{6.39/68.62}  &1.7e6/22.22 &1.7e6/22.42 \\
    GPTQ &8.13/61.09 &10.06/54.65 &6.58/60.55 &510.33/23.20 &\textbf{23.19/35.04} \\
    QuIP &10.40/58.42 &9.93/55.05 &23.09/30.82 &\textbf{254.04/23.75} &42.58/25.58 \\
    OmniQ &10.41/51.98 &14.94/40.79 &3.4e4/22.35  &852.10/22.22 &1.6e5/21.85 \\
   \bottomrule[1.2pt]
  \end{tabular}
  \label{llm3-ppl}
 \vspace{-1em}
\end{table*}

\section{PTQ-Bench: Benchmarking PTQ in LLMs}
\label{section:evaluation}

To clearly grasp the specific performance trait of each PTQ strategy and provide useful recommendations, in this section we first introduce our PTQ-Bench, then with its guidance we conduct extensive experiments to benchmark each PTQ strategy. 
%We first introduce the models, datasets, baselines and experimental settings used for evaluation. Subsequently, the detailed experimental results are listed in each subsection with corresponding analysis, conclusions and recommendations.
The detailed experimental results are listed in each subsection with corresponding analysis, conclusions and recommendations. We also list the FP16 results of all the evaluated LLMs in Table \ref{fp-llm} and \ref{fp-mllm} in Appendix \ref{fp}.

\subsection{PTQ-Bench Setup}

\paragraph{General setup}For assisting in selecting foundational PTQ strategies based on requirements, we select AWQ \citep{lin2024awq}, GPTQ \citep{frantar2022gptq}, OmniQuant \citep{shao2023omniquant}, and QuIP \citep{chee2024quip} as representatives of the four PTQ strategies, owing to their superior performance and broad practical deployment. We mainly focus on the LLaMA family (LLaMA-1/2/3/3.1) \citep{touvron2023llama1,touvron2023llama2,dubey2024llama}, the most widely deployed open-sourced LLMs. Besides, Mixtral \citep{jiang2024mixtral}, DeepSeeK-MoE \citep{dai2024deepseekmoe}, Mamba \citep{gu2023mamba}, LLaVA-1.5 \citep{liu2024visual}, and VILA-1.5 \citep{lin2023vila} are also included. The evaluation metrics contains perplexity and average inference accuracy on 16 single/multi-modal reasoning tasks. Please refer to Appendix \ref{detailed setup} for detailed experimental settings. 

\paragraph{Cross-Bitwidth Robustness} Cross-bitwidth robustness quantifies an algorithm's performance stability across varying quantization bitwidths. In PTQ-bench, we evaluate this capability by comparing the performance degradation from 4-bit to 2-bit mainly on LLaMA families.

\paragraph{Cross-structure Robustness} Cross-structure robustness evaluates an algorithm's scalability and performance consistency when deployed across diverse model structures, characterizing its generalization capability beyond specific network structures. In PTQ-bench, we compare the results on MLLMs/Mamba and LLMs with traditional structures (\textit{i.e.}, LLaMA) to achieve this goal.

\paragraph{Cross-Modality Robustness} Cross-modality robustness indicates the performance consistency of an LLM PTQ method when deployed on MLLMs. In PTQ-Bench, we mainly evaluate this capability by comparing the performance trend on LLMs and MLLMs.

\subsection{Cross-Bitwidth Robustness}
To comprehensively evaluate the cross-bitwidth robustness of the four PTQ strategies, we use LLaMA families, which are the most widely deployed open-sourced LLMs. We present the average perplexity and accuracy in Table \ref{llm-ppl} and Table \ref{llm3-ppl}. In our exploration, we observe that even under identical experimental conditions, the performance of each PTQ strategy varies significantly across bitwidths.

\paragraph{Salience-based Strategy Demonstrates Superiority at Higher-bit.}

As Table \ref{llm-ppl} and Table \ref{llm3-ppl} shown, when applying 4-bit quantization, all baselines perform comparable and satisfactory, while AWQ holds a slight advantage. When it comes to 3-bit, the performance of different baselines begins to diverge but AWQ consistently performs the best. For example, For example, on LLaMA3-8B, AWQ achieves 7.98/61.23\% and OmniQuant is 8.82/57.36\% at 4-bit. Then at 3-bit the performance of OmniQuant decline visibly with 17.53/36.09\%, but AWQ still achieves 9.83/54.65\%. \textit{\textbf{This suggests that the salience-based strategy is suitable for higher-bit PTQ}}.

\paragraph{Extremely Low-bit PTQ Performance Varies} %with Training Level of LLMs.}
According to the results in Table \ref{llm-ppl} and Table \ref{llm3-ppl}, we observe that the performance of each PTQ strategy varies significantly at 2-bit, as summarized in three dimensions below: 

\textbf{-}\textit{ Salience-based methods collapse on All LLMs:} As listed in Table \ref{llm-ppl} and Table \ref{llm3-ppl}, AWQ exhibits extremely poor performance at 2-bit across all LLMs, completely losing any language capabilities, \textit{e.g.}, on LLaMA-65B AWQ only delivers 7.4e4/22.86\%.

\textbf{-}\textit{ Optimization-based methods collapse on LLaMA-3/3.1:} Although OmniQuant performs well on LLaMA-1/2, its performance drastically declines on LLaMA-3/3.1, particularly at 3-bit where other baselines still maintain decent performance, OmniQuant has already collapsed. For instance, on 3-bit LLaMA3-70B, OmniQuant achieves 7.4e4/22.22\%, whereas GPTQ is 7.16/40.51\%.

\textbf{-}\textit{ Rotation-based and Compensation-based methods demonstrate diverse low-bit robustness on LLaMA-1/2 and LLaMA-3/3.1:} It is evident that GPTQ and QuIP consistently achieve satisfactory results across any model and bitwidth. For example, GPTQ scores 19.50/29.55\% while AWQ scores 1.1e5/22.61\% on 2-bit LLaMA2-13B, and on 2-bit LLaMA3.1-70B, QuIP is 42.58/25.58\%, whereas OmniQuant achieves 1.6e5/21.85\%. Specifically, QuIP is more suitable for LLaMA-1/2, while GPTQ demonstrates stronger robustness for 2-bit quantization of LLaMA-3/3.1. For instance, QuIP (7.70/48.47\%) outperforms GPTQ (9.00/43.44\%) on LLaMA2-70B while on LLaMA3-70B GPTQ (23.43/29.9\%) surpasses QuIP (53.18/24.60\%). To elucidate this multifaceted phenomenon, we provide the following analysis: As the technical report \citep{dubey2024llama}, the training dataset of LLaMA-3/3.1 is several times larger than that of LLaMA-1/2. The more extensive training leads to greater information loss when quantizing LLaMA-3/3.1, especially for extremely low-bit PTQ (2-bit). Our experimental results in Table \ref{llm-ppl} and Table \ref{llm3-ppl} substantiate this inference and align with the recently proposed quantization scaling law \citep{kumar2024scaling, ouyang2024low}. 

\textit{\textbf{Overall, the observation in low-bit evaluation suggests that both compensation and rotation methods are applicable for low-bit PTQ, with rotation is better suited for undertrained LLMs (LLaMA-1/2) and compensation is more advantageous for fully trained LLMs (LLaMA-3/3.1).}}

\subsection{Cross-Structure Robustness}
\label{cross-structure}

LLMs constructed by stacking traditional transformer modules suffer from high computational costs and poor long-sequence modeling capability. Consequently, various novel structures for LLMs have emerged recently, such as MoE \citep{fedus2022switch} and Mamba \citep{gu2023mamba}. However, changes in model structure may lead to potential issues of algorithm compatibility or performance degradation. Therefore, in PTQ-Bench we further evaluate the four PTQ strategies on novel-structure LLMs to analyze their cross-structure robustness. Specifically, we select Mamba-1.4B/2.8B as the representives of Mamba families, while Mixtral 8$\times$7B \citep{jiang2024mixtral} and DeepSeekMoE-16B \citep{dai2024deepseekmoe} are chosen for benchmarking MoE LLMs. The results are shown in Table \ref{mamba-moe-ppl}. For more details data please refer to Appendix \ref{detailed}.

\begin{table}[t]
  \small
  \centering
  \setlength{\tabcolsep}{5pt}
    \renewcommand{\arraystretch}{0.8} 
  \caption{The average perplexity($\downarrow$)/accuracy($\%$) comparison results on MoE and Mamba LLMs. For detailed results please refer to Table \ref{moe-pfm} and \ref{mamba-pfm} in Appendix \ref{detailed}.}
  \begin{tabular}{c|c||c|c|c}
    \toprule[1.2pt]
    \textbf{Model} & \textbf{Methods} &W4 & W3 & W2\\
     %\cmidrule(r){3-11} %\cmidrule(r){4-7} \cmidrule(l){9-12}
    \midrule %\cmidrule(r){4-7} \cmidrule(l){9-12}
    & AWQ & - & - & - \\
    Mixtral-& GPTQ &\textbf{5.65/64.59} &\textbf{6.43/59.04} &\textbf{22.92/28.19}\\
    8$\times$7B& QuIP &30.42/31.20 &39.97/28.40 &155.14/23.52\\
    & OmniQ &5.69/64.50 &6.91/56.89 &4.7e3/22.21\\
    \cmidrule{1-5}
    & AWQ & - & - & - \\
    DeepSeekMoE-& GPTQ &\textbf{7.95/53.44} &\textbf{8.75}/50.35 &49.10/27.20\\
    16B& QuIP &8.61/52.69 &8.97/\textbf{50.55} &\textbf{23.39/32.79}\\
    & OmniQ &8.21/52.36 &9.62/46.27 &75.41/25.04\\
    \midrule
    \multirow{4}{*}{Mamba-1.4B}& AWQ & - & - & - \\
    & GPTQ &\textbf{12.74/43.12}  &\textbf{15.64/40.18} &882.35/24.53  \\
    & QuIP  &14.53/41.63 &17.64/38.79 &\textbf{287.48/25.24}\\
    & OmniQ &4.2e4/25.66 &1.6e3/27.36  &2.9e4/23.05 \\
    \midrule
    \multirow{4}{*}{Mamba-2.8B}& AWQ & - & - & - \\
    & GPTQ &\textbf{11.30/45.69} &\textbf{13.35/43.84}  &696.21/24.06\\
    & QuIP &12.30/45.11  &14.15/41.97 &\textbf{119.95/28.23}\\
    & OmniQ &20.16/38.28 &36.92/34.93 &1.2e4/23.23\\
    \bottomrule[1.2pt]
  \end{tabular}
   \label{mamba-moe-ppl}
    \vspace{-1em}
\end{table}

\paragraph{Salience-based method cannot be generalized to Mamba and MoE LLMs.} AWQ requires determining the scaling hyper-parameters which must  based on the input activation during runtime. To ensure output invariance, this parameter must be integrated into the preceding linear layer. For Mamba, there is only an RMSNorm layer before $in\_proj$ layer while no linear layers precede the other projection layers, this making the output invariance cannot be ensured. For MoE LLMs, a router may choose different experts in the routing mechanism, which makes it hard to fuse the scaling hyper-parameters offline.

\paragraph{Optimization-based method is highly unstable. }
As shown in Table \ref{mamba-moe-ppl}, OmniQuant completely collapses on Mamba LLMs. Meanwhile, on Mixtral 8$\times$7B and DeepSeekMoE-16B, its performance is also unsatisfactory, particularly at 2-bit. \textit{\textbf{The aforementioned two phenomena demonstrate that AWQ and OmniQuant exhibit poor cross-structure generalization capabilities}}.

\paragraph{Rotation-based and compensation-based exhibit distinct robustness characteristics. }
The results in Table \ref{mamba-moe-ppl} indicate that GPTQ and QuIP achieve superior performance across bitwidths for both MoE LLMs and Mamba. Taking the performance on LLaMA into consideration, it is evident that these two PTQ strategies maintain stability across different model-structure. 
Interestingly, their robustness exhibit distinct nature. Specifically, in most cases, GPTQ performs better at higher bit, while QuIP is more suitable for 2-bit quantization. For example, GPTQ achieves 11.30/45.69\% on Mamba-2.8B (W4) while QuIP is 12.30/45.11\%, but at W2 QuIP outperforms GPTQ with 119.95/28.23\%. And on 2-bit DeepSeekMoE-16B, QuIP shows significant advantages with 23.29/32.79\% compared with GPTQ. In addition, we can notice that QuIP exhibits unstable performance on MoE LLMs, especially on Mixtral 8$\times$7B where GPTQ significantly outperforms QuIP across all quantization bitwidths. For instance, GPTQ and QuIP exhibit 22.92/28.19\% and 155.14/23.52\% on Mixtral at W2, respectively. \textit{\textbf{The observed phenomenon suggests that compensation-based is the most cross-structure robust strategy but rotation-based strategy excels particularly at 2-bit PTQ in most cases.} }

\subsection{Cross-Modality Robustness}
Quantization may undermine the inherent cross-modal alignment capabilities, leading to degraded performance on multimodal tasks. In PTQ-Bench, we further evaluate the four PTQ strategies in visual-language reasoning tasks on VILA \citep{lin2023vila} and LLaVA \citep{liu2024visual} to explore cross-modality robustness, and the results are presented in Table \ref{mllm-acc}. For more details please refer to Appendix \ref{detailed}.

\paragraph{Higher-bit performance remains stable and comparable. } Similar to LLMs, at 3/4-bit, the performance differences among various PTQ strategies are negligible, and all exhibit outstanding performance. For instance, on LLaVA1.5-13B the largest average accuracy gap is only 1.7\% at 3-bit.

\paragraph{Rotation and compensation methods consistently demonstrate superiority at 2-bit.} As shown in Table \ref{mllm-acc}, the phenomenon at 2-bit is consistent with that in LLMs, where only models quantized by GPTQ and QuIP exhibit effective reasoning capabilities while AWQ and OmniQuant completely collapse. For instance, on VILA1.5-13B AWQ shows 7.69\% and QuIP remains 54.18\%. \textit{\textbf{The experimental observations suggest that at higher bit, any strategy exhibits commendable cross-modal robustness, whereas at extremely low-bit,  rotation-based and compensation-based strategies emerge as the sole viable alternatives}}. %Both discoveries are almost identical to the conclusions on LLaMA.

After all evaluation, we theoretically analyze the reason why certain methods fail or work better in specific scenarios for providing deeper insights into these PTQ strategies, please refer to Appendix \ref{analysis}.

\begin{table}[t]
  \small
  \centering
  \setlength{\tabcolsep}{10pt}
  \renewcommand{\arraystretch}{0.8} 
  \caption{The average accuracy ($\%$) comparison on VILA1.5 and LLaVA1.5. For detailed results please refer to Table \ref{mllm-pfm} in Appendix \ref{detailed}.}
  \begin{tabular}{c||c|cc|cc}
    \toprule[1.2pt]
    \textbf{Methods} &W4 & \multicolumn{2}{c|}{W3} & \multicolumn{2}{c}{W2}\\
     %\cmidrule(r){3-11} %\cmidrule(r){4-7} \cmidrule(l){9-12}
    \midrule
    \textbf{VILA1.5-} &7B & 7B  & 13B & 7B & 13B \\ 
    \midrule %\cmidrule(r){4-7} \cmidrule(l){9-12}
    AWQ &68.55 &\textbf{66.30} &\textbf{69.91} &7.56 &7.69\\
    GPTQ &\textbf{68.59} &65.95 &69.33&19.30 &45.93\\
    QuIP &63.52 &64.46 &68.96 &\textbf{29.42} &\textbf{54.17}\\
    OmniQ &67.71 &65.37 &67.68 &19.99 &8.39\\
    \cmidrule{1-6}
    \textbf{LLaVA1.5-} &7B & 7B  & 13B & 7B & 13B \\ 
    \midrule %\cmidrule(r){4-7} \cmidrule(l){9-12}
    AWQ &\textbf{61.97} &\textbf{60.76} &\textbf{64.08} &7.61 &7.77\\
    GPTQ &61.74 &59.36 &63.75 &17.45 &38.46\\
    QuIP &61.32 &56.18 &64.02 &\textbf{24.80} &\textbf{45.59}\\
    OmniQ &61.27 &56.94 &62.38 &8.61 &7.80\\
    \bottomrule[1.2pt]
  \end{tabular}
  \label{mllm-acc}
\end{table}

\begin{figure}[t]
\centering
\includegraphics[width=5.5in]{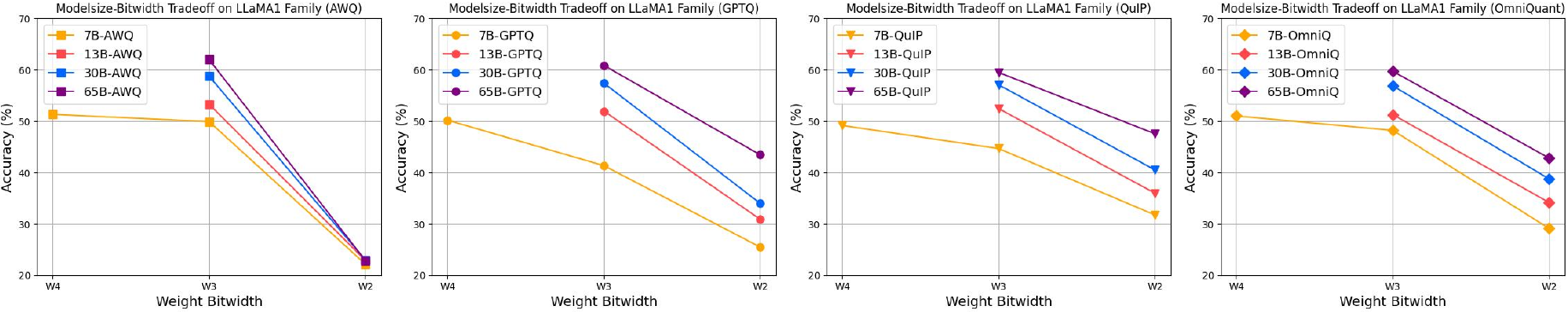}
\caption{Performance varies with model size and quantization bitwidth on LLMs. Regardless of the PTQ strategy used, the performance of a 2-bit large model is always inferior to that of a 4-bit smaller model, exemplified by 2-bit LLaMA-65B and 4-bit LLaMA-7B. In addition, 3-bit PTQ can still showcase the performance benefits associated with larger model sizes.}
\label{tradeoff}
\vspace{-0.5em}
\end{figure}

\section{Beyond Benchmarking: Extended Insights and Implications}
\label{section:analysis}

In addition to the comprehensive evaluation for the categorized PTQ strategies and recommendations in different scenarios in Section \ref{section:evaluation}, in this section we present two innovative and critical implications deriving from PTQ-Bench: a) Modelsize-bitwidth tradeoff; b) compensation is the unified robust foundational strategy.

\subsection{Trade-off among Bitwidth, Model Size and Performance}
When deploying quantized LLMs in real system, people often grapple with the decision regarding the size and the bitwidth of the model to deploy. For instance, a straightforward question arises: \textbf{\textit{Which one is better, a higher-bit smaller model or a lower-bit larger model?}} In order to clarify this confusion, we comprehensively explore the trade-off between model size and quantization bitwidth from the performance perspective. The experimental results in previous tables indicate that the perplexity of text generation exhibits a strictly positive correlation with reasoning ability reflected in accuracy. As perplexity exhibits a much larger distribution range, we choose accuracy as the evaluation metric and then create intuitive visualizations to explore the trade-off (see Figure \ref{tradeoff}).

\paragraph{The ultra-large model at 2-bit even falls short in performance when compared to the 4-bit smallest model.} The scaling law claims that, within the same LLM family, larger models generally exhibit superior performance. However, Figure \ref{tradeoff} indicates that even the largest models, like LLaMA-65B, when quantized to extremely low-bit (2-bit), demonstrate inferior performance compared to the smallest models operating at 4-bit, such as LLaMA-7B. This observation remains valid across any LLM family, any structure and any PTQ strategy as the results in the tables above. For instance, LLaMA2-70B quantized to 2-bit using OmniQuant achieves 10.36/38.46\%, which is worse than LLaMA2-7B at 4-bit with 6.55/51.11\%;
%LLaMA3-70B quantized to 2-bit using GPTQ achieves 23.19/35.04\%, which is still worse than LLaMA3-8B at 4-bit with 7.96/60.77\%;
Mamba-2.8B at 2-bit is also worse than Mamba-1.4B at 4-bit quantized by any baseline. \textit{\textbf{This fresh finding suggests that \textbf{for the present moment, deploying smaller models at higher bitwidths appears to be the optimal choice}. Additionally, specific research on extremely low-bit PTQ for ultra-large models seems essential}}. Most previous PTQ methods usually validated their performance across various model sizes, which ignored the fact that pushing ultra-large models to extremely low-bit does not surpass the performance of higher-bit smaller models so as to cause the unnecessary expenditure of computational resources. This highlights the need for the development of specialized designs for extremely low-bit PTQ on ultra-large models to ensure their performance exceeds that of higher-bit small models.

\paragraph{3-bit is an effective and competitive target for PTQ.} As illustrated in Figure \ref{tradeoff}, compared to the 4-bit smaller model, the performance advantage conferred by the large amounts of weights in the larger model remains fully evident at 3-bit, aligning with the scaling law. For example, LLaMA2-13B quantized by GPTQ exhibits 6.28/55.44\%, outperforming 4-bit LLaMA2-7B with 6.45/52.86\%. \textit{\textbf{Given that 3-bit quantization still offers a considerable target bitwidth, it can serve as a viable quantization target for those seeking to harness the performance benefits of larger models}}.

\begin{wraptable}{r}{8cm}
\vspace{-0.7em}
\setlength{\tabcolsep}{6pt}
\centering
\scriptsize
  \caption{The average perplexity($\downarrow$)/ accuracy($\%$) comparison among different PTQ strategy combination.}
  \begin{tabular}{c|c|c||cc}
    \toprule[1.5pt]
    \textbf{Model} & \textbf{Methods} & \textbf{Strategy} &PPL $\downarrow$ &Acc $\%$ \\
     %\cmidrule(r){3-11} %\cmidrule(r){4-7} \cmidrule(l){9-12}
    \midrule %\cmidrule(r){4-7} \cmidrule(l){9-12}
    \multirow{4}{*}{LLaMA} &GPTQ &Compen. &510.33 &23.20 \\
    \multirow{4}{*}{3.1-8B}&OWQ & Salien. &305.77 & 24.33\\
    & LRQuant &Optim. &193.83 &26.05\\
    & PBLLM &\textbf{Compen.}+Salien. &\underline{84.45} &\underline{33.27}\\
    & QuaRot+GPTQ &\textbf{Compen.}+Rot. &\textbf{33.33} &\textbf{35.28}\\
    \midrule
    \multirow{4}{*}{Mixtral} &GPTQ&Compen. &22.92 & 28.19\\
    \multirow{4}{*}{8$\times$7B} &OWQ & Salien. &35.60 &32.89\\
    &LRQuant &Optim. &41.81 &29.77\\
    &QuIP+GPTQ &\textbf{Compen.}+Rot. &\underline{9.59} &\underline{45.26}\\
    &PBLLM &\textbf{Compen.}+Salien. &\textbf{7.60} &\textbf{56.28}\\
    \bottomrule[1.5pt]
  \end{tabular}
  \label{gptq-base-llama}
\end{wraptable}

\subsection{Compensation-based PTQ: A Unified Robust Foundational Strategy}

In Section \ref{section:evaluation}, we have provided several recommendations for the selection of foundational PTQ strategy based on different requirements. But when developing novel PTQ algorithm, researchers usually strive for an ideal form, one where the designed method can be effectively generalized to any quantization scenario. To address this issue, in this part we further explore this valuable requirement to improve our benchmark.

Based on the results in Section \ref{section:evaluation}, it is evident that rotation-based and compensation-based strategies exhibit better generalization ability. Subsequently, we broaden our focus to more advanced algorithms within compensation-based and rotation-based strategies, \textit{i.e.}, VPTQ and QuaRot. However, when validating on Mamba and MoE LLMs, we discovered that QuaRot \citep{ashkboos2024quarot}, the state-of-the-art rotation-based method, could not be applied to Mamba and MoE LLMs, because QuaRot introduces additional rotation matrices that need to be integrated into the nearby linear layers. However, due to changes in the model structure, this integration process cannot be achieved. In contrast, any compensation-based method remains unaffected, as this strategy only relies on the weights themselves to complete error compensation. 

Subsequently, we combine GPTQ with other strategies and compare them with more advanced algorithms in other strategies to highlight the superiority in challenging scenarios including extremely low-bit (2-bit), high training level (LLaMA3.1-8B), and cross-structure (Mixtral 8$\times$7B) settings. We select naive GPTQ, OWQ (SOTA salience-based method) \citep{lee2024owq}, LRQuant (SOTA optimization-based method) \citep{zhao2024lrquant}, QuaRot+GPTQ (rotation+compensation), and PBLLM (salience+compensation) \citep{shang2023pb}. Table \ref{gptq-base-llama} reveals the PBLLM and QuaRot+GPTQ exhibit consistently high stability, outperforming more advanced salience-based and optimization-based methods as well as naive GPTQ. \textbf{\textit{Our exploration claims that compensation-based PTQ is the most unified robust strategy, and its combination with other strategies can significantly elevate the performance ceiling.}}

\section{Conclusion}
\label{section:conclusion}
Existing PTQ research lacks in-depth analysis of the superiority and applicable scenarios of various PTQ frameworks, presenting a challenge for future researchers in selecting foundational frameworks that align with their specific requirements for development. To address this issue, in this paper we first propose a comprehensive taxonomy for existing baselines. Then we establish a novel benchmark for LLMs PTQ, named PTQ-Bench, to help select foundational strategies by evaluating the cross-bitwidth, cross-structure and cross-modality robustness of the categorized strategies on a wide range of LLMs and metrics. Based on the results, we provide a series of analysis and recommendations. For example, rotation-based method shows excellent low-bit robustness. In addition, we also discover two critical insights deriving from PTQ-Bench including modelsize-bitwidth tradeoff and the most robustness strategy combination across various scenarios. We believe our PTQ-Bench will provide valuable guidelines for future development on PTQ methods.

\bibliographystyle{plainnat}
\bibliography{neurips_2025}

%%%%%%%%%%%%%%%%%%%%%%%%%%%%%%%%%%%%%%%%%%%%%%%%%%%%%%%%%%%%

\newpage
\appendix

\section*{Appendix}
\label{sec:appendix}

\section{Background}
\label{section:background}

\subsection{Quantization Preliminaries}

Quantization aims to reduce inference and storage overheads by converting high precision floating-point values into their corresponding low precision integer counterparts \citep{nagel2021white}. The asymmetric weight-only quantization is formulated as:
\begin{equation}
\mbox{W}_{q} = \mbox{clamp}(\lfloor\frac{\mbox{W}}{s_{q}}\rceil+z_{q},0,2^{b}-1),
\label{higher-bit}
\end{equation}
where $\mbox{W} \in \mathbb{R}^{n \times m}$ and $\mbox{W}_{q} \in \mathbb{R}^{n \times m}$ indicate full-precision and quantized weights respectively. $\lfloor \cdot \rceil$ denotes round-to-nearest operator. $s_{q}$ is the scaling factor and $z_{q}$ is the zero-point.

\subsection{Quantization Surveys and Benchmarks}
Most model compression reviews summarize ample concepts, principles, and a coarse-grained classification of commonly used compression methods \citep{tang2024survey,park2024comprehensive,miao2023towards,wan2023efficient,zhu2023survey, yang2024llmcbench}, but limited focus is placed on quantization. The other literature contributes a more detailed exploration of quantization. \citet{wang2024model} provides comprehensive explanations of quantization frameworks but lacks detailed taxonomy and summaries of their specific characteristics, so that the followers are still confused by how to select a basic framework to further develop. \citet{gong2024survey} delves deeper into the existing methods and provides finer distinctions but no experiments are conducted to evaluate the performance of each category. \citet{kurtic2024give} explores the accuracy-performance trade-off for popular quantization formats via broad and automated benchmarks, but only 3 models are included into examination which undermines the persuasiveness of the study. \citet{li2024evaluating} focus on synthesizing experimental results but offers limited valuable guidance for future research. It is evident that, there is still a lack of an technical and development guideline for LLMs PTQ which enables researchers to select correct model for deployment or foundational PTQ strategy based on their requirements. Our benchmark fills this gap by experimentally analyzing the characteristics of different PTQ strategies and providing recommendations for researchers to consider.

\begin{wrapfigure}{r}{0.4\textwidth}
\vspace{-1.5em}
\centering
\includegraphics[width=2.5in]{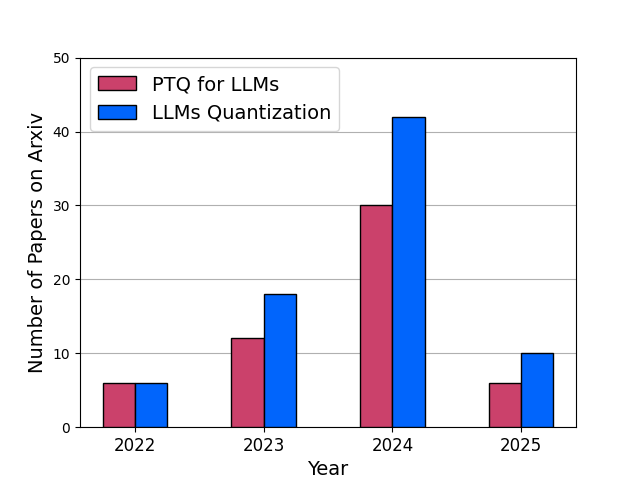}
\caption{The number of quantization papers since 2022.}
\label{trend}
\vspace{-2em}
\end{wrapfigure}

\subsection{The Trend of PTQ Research}
PTQ is more favored by LLMs compression due to its efficiency, convenience, and ease of reproducibility. As illustrated by Figure \ref{trend}, in the first year of the emergence of LLMs (2022), 6 PTQ papers were published on Arxiv. By 2023, this number doubled, while there were only 6 QAT papers. By 2024, there was an explosive growth in the number of quantization papers, with 33 out of a total of 44 papers belonging to PTQ. Overall, PTQ papers account for 69.23\% of the total number of quantization papers on LLMs, which demonstrates the intense attention on PTQ research.

\section{Details of Evaluation}
\subsection{Detailed Experimental Setup}
\label{detailed setup}
\paragraph{Baselines} AWQ \citep{lin2024awq}, GPTQ \citep{frantar2022gptq}, OmniQuant \citep{shao2023omniquant}, and QuIP \citep{chee2024quip} serve as representatives of the four PTQ strategies in our taxonomy, owing to their superior performance and broad practical deployment in numerous released LLMs. All baselines perform channel-wise quantization and set up as their description. %Consequently, our benchmark provides in-depth analyses and recommendations for the taxonomy by comparing their quantization performance across various scenarios, tasks and models. The experiments about more advanced PTQ approaches of each strategy can be found in Appendix.

\paragraph{Models} To demonstrate the universality and generalizability of our benchmark, we select LLaMA family, the most influential and widely used open-source LLM family, for performance evaluation. Specifically, this includes LLaMA-1 (7B to 70B) \citep{touvron2023llama1}, LLaMA-2 (7B to 70B) \citep{touvron2023llama2}, as well as LLaMA-3 and LLaMA-3.1 (8B and 70B)  \citep{dubey2024llama}, covering a substantial range of model sizes. Unlike other benchmarks, we further broaden our scope to investigate the cross-modality and cross-structure capabilities of different PTQ strategies. Particularly, Mixtral \citep{jiang2024mixtral}, DeepSeeK-MoE \citep{dai2024deepseekmoe} and Mamba \citep{gu2023mamba} are used to evaluate cross-structure capabilities. LLaVA-1.5 \citep{liu2024visual} and VILA-1.5 \citep{lin2023vila} are employed to assess cross-modality capabilities.

\paragraph{Datasets}
For LLMs, MoE LLMs and Mamba, we evaluate the perplexity on WikiText2 \citep{merity2016pointer} and C4 \citep{raffel2020exploring}, and zero-shot accuracies on 9 commonsense reasoning tasks, including WinoGrande \citep{sakaguchi2021winogrande}, Race \citep{lai2017race}, LAMBADA \citep{paperno2016lambada}, PIQA \citep{bisk2020piqa}, MMLU \citep{hendrycks2020measuring}, CEval \citep{huang2023ceval}, GSM8K \citep{cobbe2021gsm8k}, HellaSwag \citep{zellers2019hellaswag} and ARC \citep{clark2018think}. For MLLMs, we evaluate their visual-language reasoning abilities on VQA \citep{antol2015vqa}, GQA \citep{hudson2019gqa}, VizWiz \citep{gurari2018vizwiz}, ScienceQA \citep{lu2022learn}, TextVQA \citep{singh2019towards}, POPE \citep{li2023evaluating}, MMBench \citep{liu2025mmbench} and MMVet \citep{yu2023mm}.

\paragraph{Implementation Details}
For all evaluated models, the calibration data consists of 128 random 2048 token-segments from WikiText2. Using a batch size of 1, the entire quantization and inference processes are implemented using PyTorch \citep{paszke2019pytorch} package and deployed on 8 NVIDIA A800-80G GPUs. Codes for evaluation will be packed and open-sourced in the future.

\subsection{Detailed Evaluation Results}
\label{detailed}

In this section, we present the detailed perplexity and reasoning accuracy on specific tasks. Table \ref{llama1-pfm}, \ref{llama2-pfm}, \ref{llama3-pfm} and \ref{llama31-pfm} show the detailed results on LLaMA families (correspond to Table \ref{llm-ppl} and \ref{llm3-ppl}). Table \ref{moe-pfm} and \ref{mamba-pfm} show the detailed results on MoE and Mamba LLMs (correspond to Table \ref{mamba-moe-ppl}). Table \ref{mllm-pfm} shows the detailed results on Multimodal LLMs (correspond to Table \ref{mllm-acc}). Although our experimental evaluation is comprehensive and covers all major strategies of existing weight-only PTQ frameworks, the scale of the experiments limits us to evaluating only the most representative algorithms within each category and several latest baselines. We regard this as a limitation of this paper and will be fixed in our future work. 

\subsection{Full-precision Results of LLMs}
\label{fp}
In this section, we present the detailed full-precision perplexity and reasoning accuracy on specific tasks of all the evaluated LLMs. Table \ref{fp-llm} shows the results on LLaMA families, MoE LLMs and Mamba LLMs. Table \ref{fp-mllm} demonstrates the results on VILA1.5 and LLaVA1.5 families.

\section{Analysis on Certain Scenarios}
\label{analysis}
\subsection{Why Salience-based Strategy Collapses at 2-bit?}
AWQ employs scaling hyper-parameters to adjust salient weight channels, but the scaling hyper-parameters are too small. Given that 2-bit quantization restricts weights to only 4 discrete values, most scaled salient weights will be ultimately mapped to the same quantization grid as their unscaled counterparts, making 2-bit AWQ almost degenerate into 2-bit naive RTN (round to nearest).

\subsection{Why Optimization-based Strategy Collapses for LLaMA-3/3.1?}
Compared with other strategies, OmniQuant's optimization mechanism is particularly vulnerable to quantization error accumulation due to its reliance on quantized outputs from preceding layers as supervisory information for current-layer optimization. In fully trained LLMs like LLaMA-3/3.1, their pretrained weights carry higher information density so quantization will bring more negative impacts on them \citep{kumar2024scaling,allen2024physics}. This triggers a cascading effect: severe error accumulation degrades current-layer output validity, which in turn corrupts the quantization supervision for subsequent layers, ultimately creating a detrimental feedback loop.

\subsection{Why Compensation-based and Rotaion-based Strategies Demonstrate Robustness?}
We summarize that compensation-based method (GPTQ) and rotation-based method (QuIP) demosntrate superior robustness at cross-structure, low-bit and cross training-level scenario in Section \ref{section:evaluation}. Our analysis about this is as follows:

\paragraph{Cross-structure robustness} Unlike AWQ and OmniQuant, the pipeline of both GPTQ and QuIP don't depends on previous layers (e.g. fuse some hyper-parameters into previous layers) or quantized outputs from preceding layers, instead they make full concentration on the weights in current layer and full-precision outputs from preceding layers.

\paragraph{Low-bit and cross training-level robustness} GPTQ achieves real-time dynamic compensation during its column-wise quantization process, where the more quantization error current weight column generates, the more compensation will be added to the remaining unquantized weights. This error-adaptive mechanism significantly reduces the overall quantization error across the entire weight matrix. QuIP employs random orthogonal matrices to preprocess model weights to enhance incoherence among weights so that the mutual interference post-quantization will be reduced. In addition, this preprocessing induces more uniform magnitude distributions along quantization directions, and the improved uniformity minimizes rounding errors during quantization, ultimately mitigating the performance degradation caused by quantization.

\begin{table*}[t]
  %\scriptsize
  \centering
  \caption{Detailed performance of the four baselines on LLaMA models.}
    \resizebox{\textwidth}{!}{\begin{tabular}{c|c||ccccccccccc|cc}
    \toprule[1.5pt]
    Models & \textbf{Methods} &WinoG &Race &Lamb-o &Lamb-s &PiQA &MMLU &CEval &GSM8K &HellaS &ARC-e &ARC-c &Wiki & C4\\
    \midrule
   \multirow{4}{*}{LLaMA-7B-w4} &GPTQ 
    &68.51&39.33&67.84&61.85&77.75&33.81&26.23&6.44&55.29&73.91&41.13&6.39&7.75\\
    & OmniQ & 68.67&39.33&71.92&65.36&78.45&32.52&25.78&8.79&55.86&74.45&39.93&5.87&7.34\\
    & QuIP & 69.85&39.43&70.54&61.56&77.31&28.53&23.63&5.53&54.28&72.39&37.71&6.37&7.84\\
    & AWQ & 69.93&39.43&73.10&66.21&77.86&32.30&25.11&8.19&56.41&75.04&40.96&5.78&7.21\\
    \midrule
    \multirow{4}{*}{LLaMA-7B-w3} &GPTQ 
    &60.22&34.83&55.97&47.08&71.82&23.10&23.40&1.97&45.51&61.24&29.52&8.63&9.95\\
    & OmniQ &68.11&39.14&68.52&59.46&76.06&27.37&25.85&5.23&52.95&70.79&36.86&6.52&8.23\\
    & QuIP &64.48&36.65&57.27&52.03&74.05&27.92&23.92&2.50&49.50&68.39&34.56&7.89&9.85\\
    & AWQ &68.75&38.76&69.28&63.03&78.82&31.04&26.89&5.31&54.64&72.85&39.59&6.89&7.81\\
    \midrule
    \multirow{4}{*}{LLaMA-7B-w2} &GPTQ 
    &50.30 &27.07&5.37&5.17&57.39&25.34&25.03&0.00&29.38&32.68&22.54&27.71&44.01\\
    & OmniQ &53.75&28.42&15.51&10.27&63.33&23.80&23.63&0.00&36.50&42.38&22.78&26.66&32.88\\
    & QuIP &52.41&32.25&23.68&12.77&63.76&25.25&24.67&0.08&36.25&52.61&25.34&15.26&26.06\\
    & AWQ &49.49&22.10&0.00&0.00&52.72&22.95&23.03&0.00&25.36&25.46&22.53&2.6e5&2.9e5\\
    \midrule

    \multirow{4}{*}{LLaMA-13B-w3} &GPTQ 
    &71.11&39.33&72.64&64.54&77.80&37.24&25.78&9.55&57.48&73.65&42.32&5.63&7.16\\
    & OmniQ &69.53&39.33&71.34&60.8&77.91&35.14&23.85&10.54&57.27&75.46&42.92&5.69&7.33\\
    & QuIP &70.01&38.28&74.09&67.82&77.48&38.95&24.37&10.16&57.46&74.58&43.60 &5.67&7.17\\
    & AWQ &70.48&39.71&74.33&66.37&78.56&37.11&29.94&11.3&58.08&75.80 &44.45&5.52&7.07\\
    \midrule
    
    \multirow{4}{*}{LLaMA-13B-w2} &GPTQ 
    &53.51&30.05&20.43&17.52&63.11&23.07&25.26&0.00&38.29&45.54&23.04&7.52&15.96\\
    & OmniQ &56.20&32.73&23.40&17.85&67.14&23.57&23.03&0.76&42.06&59.47&29.52&13.36&18.89\\
    & QuIP &58.72&34.64&37.94&24.55&69.31&24.09&22.88&0.45&41.07&54.76&27.22&12.18&13.02\\
    & AWQ &48.62&22.01&0.00&0.00&53.16&26.89&26.30&0.00&25.60&26.18&22.87&2.8e5&2.3e5\\
    \midrule

    \multirow{4}{*}{LLaMA-30B-w3} &GPTQ 
    &74.27&39.81&74.42&68.64&79.76&51.60&29.49&25.55&61.41&78.58&47.78&4.89&6.50\\
    & OmniQ &75.06&39.62&73.49&67.65&79.76&50.13&26.67&26.08&60.83&78.87&47.44&4.74&6.59\\
    & QuIP &73.40&40.19&76.19&70.21&79.16&46.62&29.05&28.20&60.86&77.90&45.99&4.76&6.45\\
    & AWQ &74.66&39.33&76.46&71.14&80.20&52.73&33.28&26.38&61.69&79.80&50.34&4.61&6.35\\
    \midrule
    
    \multirow{4}{*}{LLaMA-30B-w2} &GPTQ 
    &58.33 &33.21 &36.04 &20.43 &65.07 &24.00 &22.96 &0.07 &41.11 &48.82 &23.72  &13.17 &13.58\\
    & OmniQ &58.17 &34.93 &41.88 &31.75 &70.78 &24.15 &23.03 &1.36 &43.98 &64.60 &31.74 &8.79 &14.77\\
    & QuIP &63.30 &35.69 &52.45 &36.64 &72.42 &26.22 &23.85 &0.15 &45.64 &58.80 &30.20 &9.36 &10.97\\
    & AWQ &50.43 &22.78 &0.00 &0.00 &52.72 &26.89 &26.30 &0.00 &25.43 &24.87 &23.38 &2.4e5 &2.4e5\\
    \midrule

    \multirow{4}{*}{LLaMA-65B-w3} &GPTQ 
    &76.24&41.24&77.76&73.45&80.58&56.15&32.54&38.74&62.89&79.92&49.83&4.16&6.04\\
    & OmniQ &75.37&40.48&76.4&69.34&80.3&54.88&31.05&37.15&62.86&79.17&49.83&4.09 &6.07\\
    & QuIP &75.22&42.87&76.69&72.19&80.52&53.62&31.58&31.01&61.69&79.46&49.32 &4.28 &6.16 \\
    & AWQ &75.45&42.11&78.69&74.89&80.63&57.71&36.03&41.24&63.7&80.68&50.51 &3.96 &5.92\\
    \midrule
    
    \multirow{4}{*}{LLaMA-65B-w2} &GPTQ 
    &64.88  &37.51  &59.40  &43.99  &73.50  &26.47  &23.25  &3.79  &48.12  &64.27  &32.68 &8.82 &10.23 \\
    & OmniQ &59.35  &37.61  &56.61  &39.10  &73.78  &24.39  &23.40  &2.27  &50.46  &69.74  &34.22 &7.69 &11.04 \\
    & QuIP &68.67  &38.09  &63.44  &54.22  &75.73  &31.14  &26.60  &6.90  &52.99  &69.11  &36.09 &7.19 &8.61\\
    & AWQ &51.14  &23.35  &0.00  &0.00  &53.05  &25.51  &25.56  &0.00  &25.59  &24.87  &22.44  &7.4e4 &7.5e4\\

    \bottomrule[1.5pt]
  \end{tabular}}
  \label{llama1-pfm}
\end{table*}

\begin{table*}[t]
  %\scriptsize
  \centering
  \caption{Detailed performance of the four baselines on LLaMA-2 models.}
    \resizebox{\textwidth}{!}{\begin{tabular}{c|c||ccccccccccc|cc}
    \toprule[1.5pt]
    Models & \textbf{Methods} &WinoG &Race &Lamb-o &Lamb-s &PiQA &MMLU &CEval &GSM8K &HellaS &ARC-e &ARC-c &Wiki & C4\\
    \midrule
   \multirow{4}{*}{LLaMA2-7B-w4} &GPTQ 
    &68.67 &39.90 &72.52 &66.91 &77.58 &41.55 &27.86 &12.74 &56.40 &75.00  &42.32 &5.67 &7.22\\
    & OmniQ &68.82 &40.00 &70.72 &64.08 &76.93 &32.67 &28.60  &9.93 &55.67 &74.12 &40.70 &5.74 &7.36\\
    & QuIP &68.22 &39.38 &70.43 &62.86 &76.59 &31.90 &27.45 &8.74 &55.58 &73.87 &39.78 &5.88 &7.61\\
    & AWQ &68.35 &39.71 &73.53 &67.18 &77.53 &40.86 &28.08 &13.34 &56.43 &75.84 &42.66 &5.60 &7.12\\
    \midrule
    \multirow{4}{*}{LLaMA2-7B-w3} &GPTQ 
    &65.82  &37.03  &68.54  &59.75  &76.44  &36.24  &27.79  &3.87  &52.94  &73.36  &39.76 &6.44 &7.95\\
    & OmniQ &65.51  &39.14  &65.46  &51.76  &74.65  &32.80  &26.45  &5.91  &52.41  &71.00  &38.14 &6.62 &8.62\\
    & QuIP &61.09  &31.00  &28.06  &27.44  &65.45  &22.94  &23.03  &1.44  &44.81  &56.57  &25.68 &18.66 &20.44\\
    & AWQ &68.11  &40.10  &70.33  &63.63  &76.33  &32.32  &28.53  &7.58  &54.77  &73.95  &41.89 &6.24 &7.80\\
    \midrule
    \multirow{4}{*}{LLaMA2-7B-w2} &GPTQ 
    &48.93  &26.60  &7.72  &9.06  &57.13  &22.97  &24.44  &0.00  &28.15  &32.11  &20.22 &36.77 &79.06\\
    & OmniQ &51.54  &27.37  &3.98  &1.47  &57.40  &22.95  &23.03  &0.00  &30.11  &38.89  &20.73 &37.32 &78.76\\
    & QuIP &51.07  &27.56  &11.20  &4.77  &59.25  &22.97  &23.18  &0.00  &32.85  &35.40  &20.73 &35.27 &52.22\\
    & AWQ &49.57  &23.06  &0.00  &0.00  &52.39  &25.51  &25.56  &0.00  &25.74  &24.75  &20.99 &2.2e5 &1.7e5\\
    \midrule

    \multirow{4}{*}{LLaMA2-13B-w3} &GPTQ 
    &71.90  &40.96  &74.69  &67.67  &77.91  &47.88  &31.20  &14.94  &57.74  &77.86  &47.10 &5.49 &7.06\\
    & OmniQ &69.38  &40.96  &70.64  &60.66  &77.97  &46.07  &28.53  &15.09  &57.45  &76.60  &43.34 &5.58 &7.39\\
    & QuIP &69.69  &40.67  &73.98  &65.52  &77.31  &44.92  &29.94  &16.00  &57.71  &75.38  &42.66 &5.61 &7.16\\
    & AWQ &71.74  &40.10  &75.16  &67.07  &77.20  &45.14  &32.62  &17.36  &58.57  &77.82  &45.48 &5.32 &6.95\\
    \midrule

    \multirow{4}{*}{LLaMA2-13B-w2} &GPTQ 
    &52.09  &31.48  &20.92  &13.02  &62.24  &23.00  &23.70  &0.00  &34.80  &42.59  &21.25 &20.05 &19.10\\
    & OmniQ &52.17  &30.81  &20.07  &10.17  &62.89  &22.95  &23.11  &0.00  &40.16  &48.23  &24.66 &17.22 &27.74\\
    & QuIP &55.72  &31.58  &33.86  &22.67  &65.45  &23.76  &23.03  &0.61  &39.65  &51.56  &25.85 &13.75 &14.71\\
    & AWQ &47.99  &22.39  &0.00  &0.00  &53.26  &26.89  &26.30  &0.00  &25.81  &23.04  &23.04 &1.2e5 &9.5e4\\
    \midrule

    \multirow{4}{*}{LLaMA2-70B-w3} &GPTQ 
    &76.64  &41.63  &78.96  &73.63  &80.79  &62.75  &38.56  &46.17  &63.08  &81.19  &51.71 &3.88 &5.88\\
    & OmniQ &75.22  &42.68  &78.05  &72.58  &80.30  &60.41  &37.30  &44.81  &62.40  &81.36  &53.24 &3.93 &6.06\\
    & QuIP &75.06  &40.29  &78.98  &73.51  &80.52  &60.36  &38.48  &41.39  &60.43  &79.08  &49.49 &4.28 &6.21\\
    & AWQ &75.53  &41.63  &79.49  &74.85  &81.72  &63.57  &42.87  &48.52  &63.56  &82.07  &54.01 &3.74 &5.81\\
    \midrule

    \multirow{4}{*}{LLaMA2-70B-w2} &GPTQ 
    &65.04  &36.84  &57.77  &43.57  &72.36  &32.51  &24.96  &3.41  &47.91  &62.63  &30.80 &8.38 &9.53\\
    & OmniQ &56.51  &37.61  &47.89  &25.34  &68.77  &30.35  &22.51  &2.20  &44.08  &59.64  &28.16 &8.01 &11.70\\
    & QuIP &70.88  &39.04  &65.77  &54.55  &75.63  &30.98  &23.25  &12.89  &51.64  &70.54  &37.97 &6.94 &8.46\\
    & AWQ &49.17  &22.39  &0.00  &0.00  &52.34  &24.65  &25.11  &0.00  &25.47  &25.88  &22.53 &7.2e4 &6.5e4\\

    \bottomrule[1.5pt]
  \end{tabular}}
  \label{llama2-pfm}
\end{table*}

\begin{table*}[t]
  %\scriptsize
  \centering
  \caption{Detailed performance of the four baselines on LLaMA-3 models.}
    \resizebox{\textwidth}{!}{\begin{tabular}{c|c||ccccccccccc|cc}
    \toprule[1.5pt]
    Models & \textbf{Methods} &WinoG &Race &Lamb-o &Lamb-s &PiQA &MMLU &CEval &GSM8K &HellaS &ARC-e &ARC-c &Wiki & C4\\
    \midrule
   \multirow{4}{*}{LLaMA3-8B-w4} &GPTQ 
    &72.69  &40.00  &74.54  &66.56  &79.54  &60.30  &43.98  &44.05  &59.41  &78.66  &48.72 &6.56 &9.36\\
    & OmniQ &71.43  &39.71  &69.96  &61.11  &78.40  &55.77  &40.49  &33.06  &58.27  &77.15  &45.65 &7.18 &10.46\\
    & QuIP &72.45  &40.00  &71.67  &62.90  &77.80  &56.83  &41.90  &37.98  &57.53  &79.00  &45.82 &7.16 &10.29\\
    & AWQ &73.56  &40.38  &73.90  &67.57  &79.27  &60.55  &46.66  &43.37  &59.33  &79.42  &49.57 &6.55 &9.41\\
    \midrule

    \multirow{4}{*}{LLaMA3-8B-w3} &GPTQ 
    &71.11  &39.23  &69.40  &58.24  &73.34  &52.15  &27.41  &17.89  &55.49  &71.17  &38.65 &9.39 &12.28\\
    & OmniQ &57.14  &34.45  &28.78  &20.20  &68.12  &26.20  &25.48  &1.82  &46.96  &59.68  &28.16 &14.70 &20.36\\
    & QuIP &69.61  &39.04  &67.18  &57.38  &75.79  &52.55  &35.14  &21.83  &54.96  &72.01  &39.68 &8.48 &11.70\\
    & AWQ &70.88  &39.14  &69.67  &60.74  &77.75  &49.18  &37.00  &21.53  &55.43  &75.80  &44.02 &8.16 &11.49\\
    \midrule

    \multirow{4}{*}{LLaMA3-8B-w2} &GPTQ 
    &52.80  &24.21  &0.97  &0.17  &52.01  &23.92  &24.07  &0.00  &27.10  &24.83  &20.90 &934.03 &361.33\\
    & OmniQ &50.12  &23.06  &0.02  &0.02  &54.08  &22.95  &23.48  &0.00  &26.50  &28.75  &20.39 
& 796.82 &2.4e3\\
    & QuIP &51.22  &24.02  &3.32  &1.86  &52.83  &23.19  &21.77  &0.08  &28.57  &26.64  &20.05 
&154.39 &106.25 \\
    & AWQ &49.01  &22.01  &0.00  &0.00  &52.61  &24.65  &25.11  &0.00  &25.59  &23.99  &21.84 &1.7e6 &2.2e6\\
     \midrule

    \multirow{4}{*}{LLaMA3-70B-w3} &GPTQ 
    &62.04  &23.64  &26.78  &58.55  &53.26  &71.16  &52.90  &0.00  &50.46  &26.60  &20.22 &5.88 &8.44\\
    & OmniQ &49.88  &22.78  &0.00  &0.00  &54.30  &22.96  &23.03  &0.00  &25.59  &25.55  &20.31 &4.9e4 &1.0e5\\
    & QuIP &51.54  &27.37  &10.38  &10.95  &59.96  &23.11  &23.03  &0.53  &29.65  &39.10  &18.77 &61.04 &76.16\\
    & AWQ &78.93  &42.01  &77.06  &71.65  &82.26  &72.82  &59.51  &73.46  &64.23  &84.76  &58.11 &4.69 &7.91\\
    \midrule

    \multirow{4}{*}{LLaMA3-70B-w2} &GPTQ 
    &57.30  &33.01  &25.60  &17.81  &56.04  &24.17  &23.92  &0.00  &39.90  &32.83  &18.60 &21.64 &25.21\\
    & OmniQ &49.64  &21.44  &0.00  &0.00  &52.34  &22.90  &25.63  &0.00  &25.70  &25.04  &22.61 &2.8e4 &5.5e4\\
    & QuIP &49.80  &25.55  &7.84  &9.28  &55.60  &23.00  &23.25  &0.00  &27.67  &29.92  &18.69 &51.73 &54.63\\
    & AWQ &50.28  &21.72  &0.00  &0.00  &52.18  &24.65  &25.11  &0.00  &25.58  &25.42  &23.04 &1.7e6 &1.4e6\\

    \bottomrule[1.5pt]
  \end{tabular}}
  \label{llama3-pfm}
\end{table*}

\begin{table*}[t]
  %\scriptsize
  \centering
  \caption{Detailed performance of the four baselines on LLaMA-3.1 models.}
    \resizebox{\textwidth}{!}{\begin{tabular}{c|c||ccccccccccc|cc}
    \toprule[1.5pt]
    Models & \textbf{Methods} &WinoG &Race &Lamb-o &Lamb-s &PiQA &MMLU &CEval &GSM8K &HellaS &ARC-e &ARC-c &Wiki & C4\\
    \midrule
   \multirow{4}{*}{LLaMA3.1-8B-w4} &GPTQ 
    &72.93  &39.68  &74.85  &64.69  &80.01  &61.04  &43.54  &46.05  &59.73  &79.88  &49.64 &6.70 &9.56\\
    & OmniQ &70.48  &38.47  &68.60  &60.70  &78.94  &57.83  &43.09  &35.41  &57.84  &78.87  &47.53 &7.19 &10.41\\
    & QuIP &72.69  &39.71  &67.82  &59.38  &78.89  &57.76  &43.39  &39.73  &57.76  &77.95  &47.53 &7.32 &10.40\\
    & AWQ &73.56  &39.62  &74.36  &65.81  &80.09  &60.80  &47.10  &45.34  &59.44  &80.98  &50.51 &6.66 &9.52\\
    \midrule

    \multirow{4}{*}{LLaMA3.1-8B-w3} &GPTQ 
    &73.40  &39.43  &70.15  &59.40  &72.25  &53.24  &33.21  &26.91  &55.39  &74.90  &42.85 &8.53 &11.59\\
    & OmniQ &60.14  &37.03  &35.46  &28.43  &71.27  &34.85  &25.78  &4.78  &49.78  &67.47  &33.70 &12.30 &17.57\\
    & QuIP &70.25  &37.70  &69.20  &59.50  &77.15  &53.02  &37.89  &23.73  &55.62  &77.65  &43.86 &8.36 &11.50\\
    & AWQ &71.03  &39.90  &68.83  &58.99  &77.86  &54.14  &37.22  &25.40  &55.38  &77.23  &44.54 &8.23 &11.58\\
    \midrule

    \multirow{4}{*}{LLaMA3.1-8B-w2} &GPTQ 
    &52.57  &23.25  &1.13  &0.62  &52.99  &23.89  &25.04  &0.00  &27.03  &26.81  &21.84 &731.53 &289.13\\
    & OmniQ &51.14  &23.06  &0.14  &0.04  &56.42  &22.95  &24.22  &0.00  &26.79  &28.84  &18.26 &538.42 &1.2e3\\
    & QuIP &49.80  &25.55  &3.20  &1.73  &56.26  &23.09  &25.33  &0.00  &28.54  &28.07  &19.71 &294.35 &213.72\\
    & AWQ &48.86  &21.53  &0.00  &0.00  &52.18  &24.65  &25.11  &0.00  &25.62  &24.24  &22.18 &1.6e6 &1.9e6\\
     \midrule

    \multirow{4}{*}{LLaMA3.1-70B-w3} &GPTQ 
    &77.90  &39.71  &75.86  &70.10  &69.42  &71.06  &54.46  &49.81  &63.28  &61.99  &32.42 &5.09 &8.07\\
    & OmniQ &50.99  &22.11  &0.00  &0.00  &51.63  &23.22  &24.89  &0.00  &25.74  &25.25  &22.01 &2.9e4 &3.9e4\\
    & QuIP &56.04  &27.56  &17.99  &18.67  &64.85  &24.52  &22.51  &1.29  &32.80  &50.21  &22.53 &18.99 &27.19\\
    & AWQ &77.82  &40.77  &76.93  &70.89  &81.23  &71.55  &58.25  &71.87  &63.72  &83.80  &58.02 &4.81 &7.97\\
    \midrule

    \multirow{4}{*}{LLaMA3.1-70B-w2} &GPTQ 
    &60.54  &34.16  &33.79  &25.48  &66.59  &26.32  &21.99  &0.99  &42.59  &47.01  &26.02 &19.28 &27.09\\
    & OmniQ &48.46  &20.96  &0.00  &0.00  &51.14  &23.10  &24.44  &0.00  &25.47  &25.51  &21.25 &8.1e4 &2.4e5\\
    & QuIP &49.96  &28.90  &4.95  &5.67  &58.92  &23.05  &23.03  &0.45  &30.72  &36.45  &19.28 &40.35 &44.80\\
    & AWQ &49.41  &21.44  &0.00  &0.00  &52.07  &24.65  &25.11  &0.00  &25.61  &25.46  &22.87 &1.8e6 &1.5e6\\

    \bottomrule[1.5pt]
  \end{tabular}}
  \label{llama31-pfm}
\end{table*}

\begin{table*}[t]
  %\scriptsize
  \centering
  \caption{Detailed performance of the four baselines on MoE LLMs.}
    \resizebox{\textwidth}{!}{\begin{tabular}{c|c||ccccccccccc|cc}
    \toprule[1.5pt]
    Models & \textbf{Methods} &WinoG &Race &Lamb-o &Lamb-s &PiQA &MMLU &CEval &GSM8K &HellaS &ARC-e &ARC-c &Wiki & C4\\
    
    \midrule
    \multirow{2}{*}{Mixtral-$8\times7$B}&GPTQ &75.85 &40.19 &76.36 &71.20 &81.66 &65.96 &44.73 &54.06 &62.08 &83.16 &55.20 &4.18 &7.11\\
    \multirow{2}{*}{-w4}& OmniQ &75.45 &40.38 &76.36 &70.35 &81.72 &65.46 &44.80 &53.53 &63.15 &83.33 &54.95 &4.18 &7.20\\
    & QuIP &57.46 &28.13 &13.93 &12.13 &67.25 &23.42 &22.21 &1.06 &33.47 &56.90 &27.22 &27.01 &33.83\\

    \midrule
    \multirow{2}{*}{Mixtral-$8\times7$B}&GPTQ &72.38 &39.52 &72.46 &64.23 &80.03 &59.66 &34.55 &35.63 &60.53 &80.26 &50.17 &4.99 &7.86\\
    \multirow{2}{*}{-w3}& OmniQ &71.51 &39.81 &67.28 &58.78 &81.34 &57.20 &38.04 &24.56 &58.50 &79.17 &49.57 &5.20 &8.62\\
    & QuIP &55.25 &27.66 &13.88 &13.86 &62.19 &23.10 &22.96 &0.38 &31.69 &40.74 &20.73 &37.52 &42.42\\

    \midrule
    \multirow{2}{*}{Mixtral-$8\times7$B}&GPTQ &51.85 &28.42 &13.14 &8.67 &60.34 &24.35 &26.00 &0.15 &34.02 &42.55 &20.56 &21.02 &24.81\\
    \multirow{2}{*}{-w2}& OmniQ &49.72 &22.49 &0.00 &0.00 &52.18 &22.97 &23.03 &0.00 &25.96 &26.68 &21.33 &3.4e3 &6.0e3\\
    & QuIP &48.70 &24.69 &1.82 &2.97 &55.39 &22.98 &24.37 &0.00 &27.55 &29.55 &20.73 &142.09 &168.18\\

    \midrule
    \multirow{2}{*}{DeepSeekMoE}&GPTQ &68.59 &38.85 &72.06 &67.81 &79.54 &37.12 &31.43 &14.56 &57.56 &76.26 &44.03 &6.65 &9.25\\
    \multirow{2}{*}{-16B-w4}& OmniQ &68.82 &39.14 &68.60 &62.80 &78.35 &36.45 &34.70 &12.96 &56.56 &74.71 &42.83 &6.85 &9.57\\
    & QuIP &69.53 &39.71 &69.09 &64.64 &78.94 &35.79 &35.59 &12.21 &56.31 &75.34 &42.41 &7.21 &10.01\\

    \midrule
    \multirow{2}{*}{DeepSeekMoE}&GPTQ &67.88 &38.28 &66.89 &60.80 &78.56 &33.42 &29.87 &8.87 &55.10 &73.86 &40.27 &7.35 &10.14\\
    \multirow{2}{*}{-16B-w3}& OmniQ &65.43 &35.69 &59.56 &51.99 &77.04 &24.87 &25.85 &5.23 &53.99 &72.05 &37.29 &8.01 &11.23\\
    & QuIP &68.98 &37.41 &68.66 &59.36 &77.09 &34.51 &31.35 &9.55 &54.24 &73.40 &41.47 &7.58 &10.36\\

    \midrule
    \multirow{2}{*}{DeepSeekMoE}&GPTQ &53.20 &28.61 &10.63 &1.26 &62.73 &24.75 &26.52 &0.30 &33.30 &40.28 &23.12 &55.45 &42.75\\
   \multirow{2}{*}{-16B-w2}& OmniQ &50.67 &25.45 &2.33 &2.33 &60.83 &23.11 &23.11 &0.15 &30.53 &36.15 &20.73 &72.41 &78.40\\
    & QuIP &56.83 &30.43 &32.95 &8.67 &67.08 &24.01 &24.96 &0.30 &37.91 &52.36 &25.17 &22.40 &24.38\\

    \bottomrule[1.5pt]
  \end{tabular}}
  \label{moe-pfm}
\end{table*}

\begin{table*}[t]
  %\scriptsize
  \centering
  \caption{Detailed performance of the four baselines on Mamba.}
    \resizebox{\textwidth}{!}{\begin{tabular}{c|c||ccccccccccc|cc}
    \toprule[1.5pt]
    Models & \textbf{Methods} &WinoG &Race &Lamb-o &Lamb-s &PiQA &MMLU &CEval &GSM8K &HellaS &ARC-e &ARC-c &Wiki & C4\\
    
    \midrule
    \multirow{2}{*}{Mamba-1.4B}&GPTQ &59.35 &33.30 &64.06 &55.79 &73.39 &25.24 &22.96 &2.05 &44.19 &64.35 &29.61 &11.24 &14.24\\
    \multirow{2}{*}{-w4}& OmniQ &51.93 &28.13 &2.99 &1.42 &59.19 &22.92 &26.15 &0.00 &29.08 &36.28 &24.15 &1.1e4 &7.3e4\\
    & QuIP &59.35 &32.54 &59.40 &52.01 &71.87 &25.49 &23.03 &0.45 &42.80 &62.79 &28.16 &13.05 &16.00\\

    \midrule
    \multirow{2}{*}{Mamba-1.4B}&GPTQ &57.06 &32.25 &53.66 &45.95 &71.93 &26.79 &22.96 &0.99 &41.61 &60.06 &28.67 &13.96 &17.31\\
   \multirow{2}{*}{-w3} & OmniQ &49.88 &26.79 &7.86 &9.80 &59.19 &24.90 &26.15 &0.00 &31.85 &40.19 &24.32 &1.7e3 &1.6e3\\
    & QuIP &55.56 &31.29 &51.25 &41.72 &70.29 &25.82 &23.03 &0.76 &39.80 &59.47 &27.65 &16.07 &19.20\\

    \midrule
    \multirow{2}{*}{Mamba-1.4B}&GPTQ &51.38 &24.50 &2.64 &1.11 &58.05 &23.93 &22.66 &0.00 &29.04 &35.23 &21.33 &1.2e3 &577.89\\
    \multirow{2}{*}{-w2} & OmniQ &48.78 &24.69 &0.00 &0.00 &54.24 &23.22 &25.78 &0.00 &26.33 &31.36 &19.11 &3.6e4 &2.3e4\\
    & QuIP &52.09 &24.59 &3.12 &1.46 &59.47 &24.88 &22.73 &0.00 &29.58 &35.52 &24.15 &345.15 &229.81\\

    \midrule
    \multirow{2}{*}{Mamba-2.8B}&GPTQ &64.17 &32.63 &67.53 &60.64 &74.65 &25.56 &24.74 &1.29 &48.53 &68.86 &33.96 &9.88 &12.71\\
    \multirow{2}{*}{-w4}& OmniQ &60.30 &32.92 &43.26 &40.42 &69.31 &23.58 &22.88 &0.99 &41.28 &58.96 &27.22 &18.38 &21.94\\
    & QuIP &62.51 &33.59 &68.62 &60.61 &73.50 &25.97 &24.52 &1.44 &46.57 &67.34 &31.57 &10.89 &13.70\\

    \midrule
    \multirow{2}{*}{Mamba-2.8B}&GPTQ &62.59 &33.49 &63.24 &55.00 &73.72 &25.76 &23.18 &0.76 &46.75 &65.91 &31.83 &11.92 &14.78\\
    \multirow{2}{*}{-w3}& OmniQ &56.91 &32.15 &26.59 &21.68 &68.01 &26.19 &25.04 &0.15 &40.91 &56.36 &30.29 &35.21 &38.62\\
    & QuIP &59.91 &32.15 &55.11 &49.82 &72.58 &26.21 &24.29 &0.61 &45.29 &65.74 &29.95 &12.72 &15.57\\

    \midrule
    \multirow{2}{*}{Mamba-2.8B}&GPTQ &49.25 &25.26 &0.37 &0.37 &57.34 &25.22 &23.25 &0.00 &28.45 &35.23 &19.88 &854.82 &537.60\\
    \multirow{2}{*}{-w2}& OmniQ &48.78 &22.78 &0.64 &0.29 &56.04 &24.85 &23.40 &0.00 &28.47 &29.97 &20.31 &1.7e4 &6.7e3\\
    & QuIP &53.43 &26.60 &12.17 &6.13 &61.48 &23.91 &24.15 &0.00 &32.55 &46.30 &23.81 &136.19 &103.70\\

    \bottomrule[1.5pt]
  \end{tabular}}
  \label{mamba-pfm}
\end{table*}

\begin{table*}[t]
  %\scriptsize
  \centering
  \caption{Detailed performance of the four baselines on Multimodal LLMs.}
    \resizebox{\textwidth}{!}{\begin{tabular}{c|c||cccccccccc}
    \toprule[1.5pt]
    Models & \textbf{Methods} &VQAv2 &GQA &Vizwiz &ScienceQA &TextVQA &POPE &MMBench &MMBench-cn &MM-Vet &Avg.\\
    
    \midrule
    \multirow{4}{*}{LLaVA1.5-7B-w4}&GPTQ &78.30 &61.80 &50.79 &69.25 &57.58 &86.90 &64.26 &55.41 &31.40 &61.74\\
    & OmniQ &77.80 &61.44 &49.28 &68.12 &56.30 &87.40 &62.89 &54.73 &33.50 &61.27\\
    & QuIP &78.00 &61.65 &49.77 &68.30 &57.43 &87.20 &62.47 &54.07 &33.00 &61.32 \\
    &AWQ &78.20 &61.73 &49.27 &69.75 &57.67 &87.00 &64.18 &56.70 &33.20  &61.97 \\

    \midrule
    \multirow{4}{*}{LLaVA1.5-7B-w3}&GPTQ &77.06 &60.64 &52.11 &67.72 &55.11 &84.30 &62.03 &47.25 &28.30 &59.36\\
    & OmniQ &76.21 &58.07 &52.54 &62.70 &51.98 &81.30 &59.71 &42.78 &27.20 &56.94 \\
    & QuIP &74.50 &57.51 &53.07 &62.61 &49.71 &87.50 &46.31 &44.60 &29.80 &56.18 \\
    &AWQ &77.43 &60.48 &53.61 &66.73 &55.97 &85.80 &64.95 &52.66 &29.20 &60.76\\

    \midrule
    \multirow{4}{*}{LLaVA1.5-7B-w2}&GPTQ &41.89 &24.65 &0.08 &1.96 &13.65 &68.60 &0.95 &0.69 &4.60 &17.45\\
    & OmniQ &0.11 &0.00  &0.00  &0.75  &0.87  &68.00  &1.29  &1.63  &4.80 &8.61\\
    & QuIP &47.26 &30.83  &19.76  &20.77  &18.45  &71.40  &4.81  &1.80  &8.10 &24.80\\
    &AWQ &0.00 &0.00  &0.00  &0.00  &0.00  &68.00  &0.00  &0.00  &0.50 &7.61\\

    \midrule
    \multirow{4}{*}{LLaVA1.5-13B-w3}&GPTQ &79.50 &62.70  &48.67  &72.46  &59.68  &87.00  &67.87  &60.91  &35.00 &63.75\\
    & OmniQ &78.80 &62.01  &50.82  &70.55  &58.23  &89.20  &64.52  &54.73  &32.60 &62.38\\
    & QuIP &79.20 &62.60  &54.45  &71.87  &59.70  &87.90  &67.87  &58.85  &33.70 &64.02\\
    &AWQ &78.80 &62.69  &52.99  &71.78  &59.84  &88.50  &67.78  &60.05  &34.30  &64.08\\

    \midrule
    \multirow{4}{*}{LLaVA1.5-13B-w2}&GPTQ &67.15 &47.52  &37.27  &26.88  &35.98  &88.20  &23.54  &2.66  &16.90 &38.46\\
    & OmniQ &0.01 &0.00  &0.00  &0.00  &0.00  &68.00  &1.46  &0.77  &0.00 &7.80\\
    & QuIP &72.40 &56.25  &50.30  &38.65  &41.93  &76.00  &44.76  &11.25  &18.80 &45.59\\
    &AWQ &0.00 &0.00  &0.00  &0.00  &0.00  &68.00  &0.60  &0.43  &0.90 &7.77\\

    \midrule
    \multirow{4}{*}{VILA1.5-7B-w4}&GPTQ &82.18 &63.00 &64.22 &78.76 &66.01 &87.50 &71.39 &63.32 &40.90 &68.59\\
    & OmniQ &81.91 &62.95 &62.80 &78.14 &65.69 &86.80 &70.96 &61.77 &38.40 &67.71\\
    & QuIP &81.22 &61.53 &56.56 &73.38 &63.09 &87.40 &66.15 &45.96 &36.40 &63.52\\
    &AWQ &82.17 &63.26 &63.87 &79.01 &65.95 &87.80 &71.65 &63.23 &40.00 &68.55\\

    \midrule
    \multirow{4}{*}{VILA1.5-7B-w3}&GPTQ &81.10 &60.93 &61.53 &76.35 &64.48 &83.90 &68.56 &60.22 &36.50 &65.95\\
    & OmniQ &80.88 &61.58 &59.83 &74.53 &63.27 &86.10 &67.01 &56.62 &38.50 &65.37\\
    & QuIP &77.11 &53.50 &47.34 &51.29 &53.35 &81.40 &45.19 &5.93&36.00 &50.12\\
    &AWQ &81.38 &62.09 &62.06 &76.75 &64.06 &85.80 &68.04 &58.33 &38.20 &66.30\\

    \midrule
    \multirow{4}{*}{VILA1.5-7B-w2}&GPTQ &32.81 &18.66 &4.32 &28.20 &13.32 &66.10 &2.23 &1.46 &6.60 &19.30\\
    & OmniQ &40.97 &15.88 &9.51 &2.00 &16.09 &74.90 &4.98 &0.34 &15.20 &19.99\\
    & QuIP &60.13 &38.20 &23.80 &25.47 &20.48 &79.20 &0.86 &0.17 &16.50 &29.42\\
    &AWQ &0.00 &0.00 &0.00 &0.00 &0.00 &68.00 &0.00 &0.00 &0.00 &7.56\\
    
    \midrule
    \multirow{4}{*}{VILA1.5-13B-w3}&GPTQ &82.30 &63.98  &62.32  &81.58  &65.17  &86.80  &74.31  &63.66  &43.90 &69.33\\
    & OmniQ &80.60 &60.35  &62.37  &78.07  &65.00  &89.90  &70.88  &61.94  &40.00 &67.68\\
    & QuIP &82.20 &62.98  &63.44  &80.92  &66.12  &86.90  &73.02  &63.23  &41.80 &68.96\\
    &AWQ &82.30 &63.17  &61.75  &81.75  &66.29  &89.20  &73.97  &66.24  &44.50 &69.91\\

    \midrule
    \multirow{4}{*}{VILA1.5-13B-w2}&GPTQ &72.70 &51.34  &45.70  &49.78  &44.54  &80.30  &38.66  &2.66  &27.70 &45.93\\
    & OmniQ &0.15 &0.00  &0.10  &0.21  &1.75  &67.90  &1.03  &0.26  &4.10 &8.39\\
    & QuIP &76.30 &55.70  &46.53  &63.17  &50.16  &82.60  &56.01  &25.43  &31.70 &54.18\\
    &AWQ &0.00 &0.00  &0.00  &0.00  &0.00  &68.00  &0.52  &0.69  &0.00 &7.69\\

    \bottomrule[1.5pt]
  \end{tabular}}
  \label{mllm-pfm}
\end{table*}

\begin{table*}[t]
  %\scriptsize
  \centering
  \caption{Detailed FP16 performance of all evaluated LLMs, MoE LLMs and Mamba LLMs.}
    \resizebox{\textwidth}{!}{\begin{tabular}{c||ccccccccccc|cc}
    \toprule[1.5pt]
    Models &WinoG &Race &Lamb-o &Lamb-s &PiQA &MMLU &CEval &GSM8K &HellaS &ARC-e &ARC-c &Wiki & C4\\
    \midrule
   LLaMA-7B&70.01 &40.29 &73.57 &67.82 &78.67 &32.18 &26.89 &9.02 &56.99 &75.29 &41.81&5.68&7.08\\
   LLaMA-13B& 72.77 & 39.62 & 76.15 & 71.08 & 79.16 & 43.48 & 24.44 & 17.13 & 59.92 & 77.36 & 46.42 & 5.09 & 6.61\\
    LLaMA-30B& 75.69 & 40.57 & 77.59 & 73.34 & 80.96 & 54.63 & 32.10 & 33.06 & 63.34 & 80.39 & 52.82 & 4.10 & 5.98\\
    LLaMA-65B& 77.35 & 41.53 & 79.10 & 74.89 & 81.34 & 59.37 & 38.93 & 46.85 & 64.57 & 81.36 & 52.82 & 3.53 & 5.62\\ 
    \midrule
    LLaMA2-7B& 69.06 & 39.52 & 73.86 & 68.23 & 78.07 & 41.84 & 29.87 & 13.57 & 57.14 & 76.30 & 43.34 & 5.47 & 6.97\\ 
    LLaMA2-13B& 72.14 & 40.57 & 76.77 & 70.33 & 79.11 & 52.10 & 36.85 & 23.12 & 60.04 & 79.46 & 48.46 & 4.88 & 6.47\\
    LLaMA2-70B& 77.90 & 42.58 & 79.58 & 74.69 & 82.21 & 65.46 & 44.28 & 53.30 & 64.79 & 82.70 & 54.35 & 3.31 & 5.52\\ 
    \midrule
    LLaMA3-8B& 73.24 & 40.29 & 75.65 & 68.72 & 79.54 & 62.23 & 48.44 & 50.87 & 60.14 & 80.09 & 50.17 & 6.14 & 8.88\\
    LLaMA3-70B& 80.66 & 42.01 & 79.33 & 73.94 & 82.32 & 75.20 & 64.56 & 80.14 & 66.29 & 86.78 & 60.41 & 2.86 & 6.73\\
    LLaMA3.1-8B& 73.95 & 39.04 & 75.45 & 67.13 & 80.14 & 63.35 & 49.03 & 49.28 & 59.99 & 81.40 & 51.54 & 6.27 & 8.99\\ 
    LLaMAa3.1-70B& 79.72 & 40.19 & 78.75 & 73.86 & 82.86 & 75.32 & 63.37 & 80.52 & 66.44 & 87.21 & 60.92 & 2.83 & 6.70\\
    \midrule
    Mixtral-8x7B& 76.48 & 39.90 & 78.01 & 73.10 & 82.43 & 68.00 & 48.66 & 58.76 & 64.79 & 84.01 & 56.74 & 3.84 & 6.88\\ 
    DeepSeekMoE-16B& 70.09 & 39.23 & 72.97 & 69.36 & 78.89 & 37.79 & 35.81 & 16.45 & 58.04 & 76.05 & 44.54 & 6.51 & 9.05\\
    \midrule
    Mamba-130m& 52.49 & 27.37 & 44.23 & 33.86 & 64.53 & 22.67 & 23.03 & 0.76 & 30.81 & 47.98 & 19.80 & 20.66 & 22.82\\
    Mamba-370m& 55.17 & 29.67 & 55.58 & 47.25 & 69.48 & 22.89 & 22.59 & 0.61 & 37.19 & 55.05 & 24.91 & 14.34 & 17.28\\ 
    Mamba-790m& 56.12 & 33.01 & 61.42 & 54.18 & 72.09 & 23.77 & 23.85 & 1.29 & 42.31 & 61.15 & 26.45 & 12.04 & 14.94\\
    Mamba-1.4B& 61.40 & 34.35 & 64.91 & 56.72 & 74.10 & 24.87 & 22.88 & 0.99 & 45.03 & 65.49 & 29.78 & 10.76 & 13.61\\
    Mamba-2.8B& 63.14 & 33.78 & 69.09 & 61.32 & 75.19 & 26.45 & 23.77 & 1.74 & 49.49 & 69.78 & 34.39 & 9.46 & 12.28\\ 
    \bottomrule[1.5pt]
  \end{tabular}}
  \label{fp-llm}
\end{table*}

\begin{table*}[t]
  %\scriptsize
  \centering
  \caption{Detailed FP16 performance of all evaluated MLLMs.}
    \resizebox{\textwidth}{!}{\begin{tabular}{c||ccccccccccc}
    \toprule[1.5pt]
    Models&VQAv2 &GQA &Vizwiz &ScienceQA &TextVQA &POPE-r &POPE-r &POPE-a &MMBench &MMBench-cn &MM-Vet\\
    \midrule
    LLaVA1.5-7B& 78.52 & 61.94 & 50.06 & 70.22 & 58.21 & 87.30 & 86.10 & 84.20 & 64.69 & 58.08 & 30.90\\
    LLaVA1.5-13B& 80.00 & 63.25 & 53.61 & 74.94 & 61.19 & 87.10 & 86.30 & 84.50 & 68.47 & 63.49 & 36.60\\ 
    \midrule
    VILA1.5-7B& 82.31 & 63.39 & 63.91 & 79.42 & 66.10 & 87.50 & 86.30 & 84.90 & 72.16 & 63.92 & 40.30\\ 
    VILA1.5-13B& 82.80 & 64.38 & 62.67 & 83.38 & 64.97 & 87.50 & 86.20 & 85.20 & 75.00 & 66.32 & 44.20\\ 
    \bottomrule[1.5pt]
  \end{tabular}}
  \label{fp-mllm}
\end{table*}

%%%%%%%%%%%%%%%%%%%%%%%%%%%%%%%%%%%%%%%%%%%%%%%%%%%%%%%%%%%%

\end{document}